\documentclass{clv3}
\usepackage[utf8]{inputenc}
\usepackage{hyperref}
\usepackage{xcolor}
\definecolor{darkblue}{rgb}{0, 0, 0.5}
\hypersetup{colorlinks=true,citecolor=darkblue, linkcolor=darkblue, urlcolor=darkblue}
\usepackage{booktabs}
\usepackage{tabularx}
\usepackage{algorithmic}
\usepackage[normalem]{ulem}

\issue{01}{01}{2021}

\dochead{Computational Linguistics Submission}

\runningtitle{Computational Linguistics}

\runningauthor{Richard Plant}

\begin{document}

\title{You Are What You Write: Preserving Privacy in the Era of Large Language Models}

\author{Richard Plant, Valerio Giuffrida and Dimitra Gkatzia}
\affil{Edinburgh Napier University\\\texttt{\{r.plant,v.giuffrida,d.gkatzia\}@napier.ac.uk}}

\maketitle

\begin{abstract}
Large scale adoption of large language models has introduced a new era of convenient knowledge transfer for a slew of natural language processing tasks. However, these models also run the risk of undermining user trust by exposing unwanted information about the data subjects, which may be extracted by a malicious party, e.g. through adversarial attacks. We present an empirical investigation into the extent of the personal information encoded into pre-trained representations by a range of popular models, and we show a positive correlation between the complexity of a model, the amount of data used in pre-training, and data leakage. In this paper, we present the first wide coverage evaluation and comparison of some of the most popular privacy-preserving algorithms, on a large, multi-lingual dataset on sentiment analysis annotated with demographic information (location, age and gender). The results show since larger and more complex models are more prone to leaking private information, use of privacy-preserving methods is highly desirable. We also find that highly privacy-preserving technologies like differential privacy (DP) can have serious model utility effects, which can be ameliorated using hybrid or metric-DP techniques.
\end{abstract}

\section{Introduction}

Natural language communication exhibits the possibility of both increased human understanding and conflict, something we intuitively demonstrate when we filter the sharing of sensitive personal information based on context. We would not reveal the same level of personal information to a casual acquaintance that we do to our doctor, since we expect a different level of confidentiality from those contexts. It is incumbent on the creators of automated systems to deal with the human norms around privacy, ensuring that we accord with social expectations around the risk of personal or reputational damage \cite{kozyreva_public_2021}.

Many recent advances in the field of natural language processing (NLP) have been driven by the introduction of large language models (LLMs) which can extend into the billions of parameters, pre-trained on very large textual corpora \cite{narayanan_scaling_2021}. These models have demonstrated state-of-the-art performance in a number of tasks and benchmarks \cite{han_pre-trained_2021}, and enabled the transferring of the embedded knowledge of such models to researchers and organisations that lack the resources to train an equivalently-complex model of their own \cite{qi_when_2018, peters_tune_2019}. However, the popularity of LLMs driven by these factors has threatened to transgress the expectation of privacy around automated text processing.  

Language models pose a threat to personal privacy because of their immense potential for misuse - text authorship can reveal a vast amount of personal information about an individual, including potentially sensitive demography revealed without the authors' knowledge or intention \cite{Rao2000pseudo,emmeryAdversarialStylometryWild2021}. Users may be less inclined to allow their writing to be used in machine learning systems if they knew that could expose their political ideology \cite{iyyer_political_2014,colleoni_echo_2014}, economic status \cite{aletras_predicting_2018,doi_estimation_2020}, or a range of other things that could uniquely identify them \cite{narayanan_robust_2008,sun_identity_2012}. Much recent research has provided evidence that such models are prone to this kind of information leakage, due to the memorization of large parts of their training corpus \cite{carlini_secret_2019,leino_stolen_2020} and the exposure of sensitive attributes through the embedding generation process \cite{Song2020}.

These risks have led to a rise in public distrust of automated systems---for an instructive example, see the persistence of the belief that Facebook or other social networks are using smartphone microphones to eavesdrop on users, despite the near-total lack of evidence \cite{kroger_is_2019}. The onus in these case is on machine learning practitioners to reassure potential users on the security and privacy of their data, or risk undermining the public trust on which we rely \cite{kozyreva_public_2021, prabhumoyeCaseStudyDeontological2021}. This principle becomes only more critical when viewed in the light of under-resourced language users, who already deal with exclusion and bias \cite{benjamin_hard_2018,nekoto_participatory_2020}.

There are however promising developments in the field of statistical privacy that present an opportunity to continue to make progress in NLP, while preserving a measurable level of privacy for the individuals whose data we rely on \cite{dwork_algorithmic_2013}. These innovations arrive at a critical time, since as we show, there are good reasons for concern over the amount of personal and private information the most effective learning tools are already leaking. We present a critical evaluation of these developments and their impact on the demographic privacy risk specific to LLMs, as well as proposing hybrid solutions that maximise the privacy provided by model-agnostic interventions.

\subsection{Privacy at risk}
\label{sec:privacy_risk}

Why should we concern ourselves with the privacy risk to individuals whose data we use in the course of machine learning research? Beyond the obvious moral objection that we should strive to minimise any potential negative impact from our work on any participant, the key role of large-scale public datasets in driving advances in NLP means that public trust is essential for continued progress in the field; dealing with the safety and security of personal data is a necessary step towards establishing that trust \cite{anwar_supporting_2021, shadbolt_privacy_2019, horvitz_data_2015}. Further, we should take some responsibility for mitigating genuine potential harms that may stem from downstream use of our research outputs: machine learning is seeing extensive use in the insurance industry \cite{paruchuri_impact_2020, baudry_machine_2019, roy_detecting_2017} for example, and it does not stretch the bounds of credulity a great deal to imagine an unscrupulous provider eliciting pieces of demographic information in this roundabout way that it would be illegal in certain jurisdictions to simply ask for, since it would provide a basis for making biased decisions about pricing and coverage.

Use of freely available mass published data, like that harvested from social media, presents similar ethical considerations \cite{townsend_social_2016}. We have a duty to ensure both consistency and privacy here: users may have edited or deleted posts that models continue to rely on in existing datasets, and may unintentionally reveal information they would retroactively choose to keep private \citep{Bartunov2012, Pontes2012, Goga2013}. These principles form an important part of the General Data Protection Regulations \cite{cummings_role_2018, hijmans_ethical_2018, kuner_machine_2017}, which imposes a statutory duty on researchers to consider these factors during experimental design. The severity of this risk, or more accurately the tolerance of a user towards disclosure and their willingness to share potentially private information, may also vary dependent on the cultural mores of their society, ethnic group, and age cohort \cite{fietkiewicz_investigating_2018, callanan_user_2016, krasnova_privacy_2010}, all of which would prove difficult, if not impossible, for researchers to control for at the instance level. There is persuasive evidence that this differential expectation of privacy risk may partially determine mores around personal information disclosure at the national level, leading to uneven levels of data availability and consistency across cultural boundaries \cite{pomfret_beyond_2020, reed_thumbs_2016}.

The massive availability of data sourced from millions of individuals across the Internet has provoked large advances in our ability to design capable algorithmic systems for modelling language tasks \cite{vaswani_attention_2017, devlin_bert_2019}. However, these advances have exposed the issue of privacy in a more urgent way than we have seen previously - the more we learn about language from texts authored by real people \cite{petroni_language_2019} and the better our models become at generalising beyond narrow task-specific use cases \cite{brown_language_2020}, the more we unavoidably learn about the people who wrote them. Indeed, a growing body of experimental literature has established that these more powerful models can leak substantial information that we may wish to keep private - personally identifying demographic information among them \citep{song_machine_2017, carlini_secret_2019}. 

Even seemingly innocuous sets of data \cite{Xu2008}, such as the network properties of social media users \cite{Pontes2012, aletras_predicting_2018} or data which has been cleansed of identifying attributes \cite{sun_identity_2012}, can provide fertile ground for the kind of attack known as re-identification \cite{frankowski_you_2006}, in which a malicious actor attempts to discover demographic characteristics---and ultimately the identity---of a target from a seemingly 'anonymous' set of data about them. It has also been established conclusively that it is practical to extract details of an individuals' presence or absence in a model's training set, in an attack known as membership inference \citep{leino_stolen_2020, truex_demystifying_2019}. Potentially sensitive information such as age or gender can be reliably reconstructed given the output of such a model \cite{zhang_locmia_2020, fredrikson_model_2015}, along with more esoteric topics such as political and ideological orientation \cite{khatua_predicting_2020, colleoni_echo_2014, iyyer_political_2014}.

Latent representations of highly complex data, including neural embeddings for text sequences, have proven to be ideal vectors for this kind of privacy vulnerability \cite{Song2020}, and thus form an important test case for determining the effectiveness of any prospective privacy-preserving intervention.

Privacy differentials across multilingual representations is a relatively under-studied phenomenon, but we consider the potential existence of differences germane, and hence we include monolingual models for languages other than English, as well as multilingual models, in our analysis. There is some evidence that multilinguality can benefit task performance when applied to low or medium resource languages \cite{priban_are_2021, pires_how_2019}, however this is contested by other studies which claim a performance advantage for monolingual models \cite{ronnqvist_is_2019, virtanen_multilingual_2019}. One possible explanation for this difference in findings is that the level of relatedness between high and low resource languages incorporated into a multilingual embedding may be determinative \cite{woller_not_2021}; if the lower-resourced language input is less related to the handful of high resource languages typically extant in language model training sets \cite{joshi_state_2020}, then the more complex multilingual model may under-perform a smaller monolingual system. We wish to extend this analysis to consider the privacy implications of both strategies.

\subsection{Existing privacy strategies}

The existence of targeted attacks against the personal privacy of individuals whose data is contained in large datasets has not led to a curtailment in the growth of available data, but to calls for the establishment of participant trust through strong privacy-preserving practices and norms \cite{daries_privacy_2014, paullada_data_2021}.

Anonymisation, or as it might be better termed pseudonymisation \cite{Rao2000pseudo, pfitzmann_anonymity_2001}, is one potential response to this trend. Under this rubric, specific identifiers that can uniquely unmask an individual are removed---a national identification or social security number for instance---and replaced with synthetic identifiers created for this experiment. Unfortunately, a string of experimental results \cite{narayanan_-anonymizing_2009, Jawurek2011deanon, biryukov_deanonymisation_2014, pyrgelis_there_2019} as well as the case of the Netflix Prize Dataset \cite{narayanan_how_2006}, in which researchers were able to uniquely identify the majority of 'anonymous' participants through reference to other public sources of information, proved that maintaining personal privacy is harder than it might first appear; that in fact, our preferences, interactions, and the data we generate may be enough in themselves to reveal parts of our identity that we would prefer remain hidden.

Pseudonymisation relies on the logically spurious assumption that the protected data will always be directly referred to in the text, and cannot be inferred from otherwise innocuous text sequences with the application of other sources of knowledge. Moreover, the cost of more advanced forms of pseudonymisation such as $k$-anonymity or $l$-diversity that attempt to provide a statistical guarantee of privacy disclosure can prove severe, to the extent that achieving a minimal level of privacy gain can almost completely eliminate the utility of a dataset for the intended purpose \cite{brickell_cost_2008, Li2009obfuscation}.

Hence, developing privacy-preserving approaches that can provide a verifiable level of privacy, while also preserving a meaningful level of performance in a given set of downstream tasks.

\subsection{Adversarial and differential privacy}

To reduce the salience of such privacy attacks, recent research has adopted the adversarial training methods pioneered in the field of domain adaptation \cite{ganin_domain-adversarial_2016}. Essentially, this strategy attempts to maximise the performance of a model when train and test instances are drawn from similar but distinct distribution, by learning an intermediate representation that promotes features that benefit the target task, while suppressing features that are heavily conditioned by the domain shift. Applying this model to privacy in NLP tasks is a short conceptual step: instead of domain-invariant features, we seek to surface private-attribute invariance while learning a good representation for our downstream task \cite{elazar_adversarial_2018}. 

To provide a quantifiable guarantee of disclosure risk, we employ the concept of differential privacy (DP) \cite{dwork_algorithmic_2013}. Under the formulation of this system (see Equation \ref{def:epsilon_privacy}), we call a computation differentially private if the results on a dataset $A$ are equally plausible as the results on a dataset $B$ that differs in a single record, with the addition of a small amount of calibrated noise. That maximum possible deviation, expressed by the parameter $\epsilon$, stands in for our maximal privacy loss.

The level of private information leaked by a computation $M$ can be expressed by the variable $\epsilon$, where for the symmetric difference $\oplus$ between two datasets $A$ and $B$ that differ in only one record, and any set of possible outputs $S \subset Range(M)$,

\begin{equation}
\label{def:epsilon_privacy}
Pr[M(A) \in S] \leq Pr[M(B) \in S] \times exp(\epsilon \times |A \oplus B|)
\end{equation}

We can apply this to our problem space by adopting the local differential privacy approach \cite{cormode_privacy_2018, MahawagaArachchige2020local}. Under this scheme, we apply noise proportional to the sensitivity of our computation - the maximum difference in output of the same operation carried out on both datasets - and the $\epsilon$ variable to the input before we carry out learning through gradient descent; we are essentially reducing the certainty of our model about the state of any individual record, and hence the ability of adversaries to carry out re-identification with intermediate representations that follow this step. 

Given a scenario in which we are aware of categories of private information that may interest an attacker such as gender, location or age, it is intuitively compelling to combine a general privacy approach such as $\epsilon$-DP along with an adversarial objective that targets the specific variables we wish to obscure. In addition to the formal guarantees provided by our DP implementation, this allows us to compare outcomes for the specific attacker task envisioned as a practical measure of privacy.

\subsection{Contributions}

In this work, we present an initial contribution towards a comparison of several promising methods for reducing the amount of private demographic information leakage during a representative NLP task. In particular:

\begin{itemize}
    \item We carry out a systematic evaluation of both adversarial training methods (gradient reversal and cross-gradient training), along with methods based on the application of differential privacy. Both paradigms represent active areas of research for NLP privacy.
    \item We present a comparison of the impact of such privatising systems across a set of input embeddings drawn from popular large pre-trained language models chosen to reflect popular models in use for knowledge transfer across the NLP domain, and show a reduction of up to $40\%$ in relative attack efficacy. 
    \item We evaluate and demonstrate our results across a multilingual corpus, demonstrating the effect of these technologies across a set of relatively high (English), medium (French and German), and lower resource (Norwegian, Danish) European languages, as well as a multilingual slice composed of records from all included languages. We note both commonalities and distinctions in our results which may present useful areas for further linguistic study.
    \item We study the impact of the privacy/utility trade-off and present an empirical basis for gauging an appropriate balance of outcomes for hybrid adversarial/differentially private systems.
\end{itemize}

On the basis of these studies we present an adversarial/differentially private hybrid model as the preferred choice for achieving relatively high levels of individual privacy, while offering a significant level of calibration sensitivity so that downstream users can rapidly achieve an acceptable privacy/utility balance. On this basis, we investigate the impact of multiple task-related factors such as input dimensionality and model complexity on optimal privacy system design.

This paper will first lay out a review of related work in the field in Section \ref{sec:related}, including some proposed methods for privatising of texts in NLP. Then, we will lay out our methodology in Section \ref{sec:methodology}, before discussing our suite of models studied, evaluation criteria, and the results in Section \ref{sec:results}. Section \ref{sec:discussion} will present some further analysis and discussion of secondary considerations arising from our experimental work, including investigation of the tradeoff between privacy and model utility. We will make some closing remarks in Section \ref{sec:conclusion}, along with some recommendations for further work.

\section{Related Work}
\label{sec:related}

\paragraph{Adversarial Learning} Adversarial training has seen strong research interest since initial proposition, with multiple works extending the paradigm
\cite{zhang_recent_2020, wang_transferable_2019, wang_deep_2018}. \citet{coavoux_privacy-preserving_2018} provided a useful guide to applying this work to personal privacy in the setting they describe as 'multi-detasking', in which we train an adversary network using intermediate representations from our main task model to recover demographic labels about the input.

\citet{Li2018adversarial} applied this method to multiple protected attributes simultaneously across several English-language datasets, an approach that was also followed by \citet{madras_learning_2018} and \citet{elazar_adversarial_2018}. \citet{Xu2019} proposed a counter-intuitive model wherein input texts are rewritten via back-translation from adversarially-trained representations to eliminate sensitive attributes from the dataset; while this may prove a vital technology for dataset owners to apply before release, we concern ourselves in this work with the intermediate representations themselves.

This form of adversarial training has also been applied to debiasing representations; in \citet{kaneko_gender-preserving_2019}, \citet{zhang_mitigating_2018}, and \citet{zhao_learning_2018} the goal is to generate gender-invariant representations that do not express a differential between binary genders. \citet{jaiswal_privacy_2020} and \citet{li_speaker-invariant_2020} extend the problem space to include multi-modal data, an interesting research direction that may show potential for generalisation beyond NLP.

\paragraph{Differential Privacy in NLP} Local differential privacy (LDP), in terms of applying additive noise to inputs before passing them through the intended computation, has seen a great deal of interest in privacy-preserving data collection \cite{erlingsson_rappor_2014, kairouz_discrete_2016, wang_using_2016}. However, previous research into LDP has also borne out the potential for this method to reduce information leakage from trained text representations \cite{beigi_privacy_2019}. \citet{lyu_differentially_2020} propose a system that adds perturbation to a pre-trained embedding vector by normalizing the vector \cite{shokri_privacy-preserving_2015}, then applying additive noise drawn from the Laplace distribution in the traditional DP fashion \cite{dwork_calibrating_2006}.

\begin{figure}[ht]
    \centering
    \includegraphics[width=0.6\textwidth]{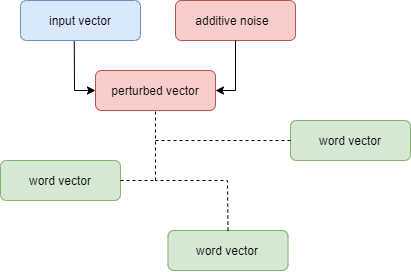}
    \caption{Metric differential privacy mechanism. Dashed lines represent an arbitrary distance calculation.}
    \label{fig:mdp}
\end{figure}

Metric differential privacy (MDP) \cite{chatzikokolakis_broadening_2013}, which we consider a corollary development to local DP, involves the substitution of the original formulation's Hamming distance metric for understanding indistinguishability of two datasets for an arbitrary distance mechanism. \citet{feyisetan_privacy-_2020} and \citet{fernandes_generalised_2019} applied this formulation to perturbing texts by proposing a system which swaps words for perturbed versions by considering the distance between their embedding vector with additive noise applied and the other vectors in the vocabulary, choosing the nearest candidate based on their metric of interest, as shown in Figure \ref{fig:mdp}. Beyond simple distance metrics, some research has also pointed to the potential value of alternate metric spaces \cite{feyisetan_leveraging_2019, xu_differentially_2020, dhingra_embedding_2018} in maintaining semantic continuity while ensuring DP-compliance.

\paragraph{Hybrid Approaches} Some previous work has married both approaches: \citet{phan_scalable_2019} proposed an approach which implements classical differential privacy in an adversarial learning paradigm, however, this work relies on adversarial objectives to promote robustness to adversarial samples rather than privacy. \citet{alnasser_privacy_2021} proposes a similar hybrid approach, applying both adversarial and local differential privacy to English language embeddings, but only for the pre-trained BERT language model. In our paper CAPE: Context-Aware Private Embeddings for Private Language Learning \cite{plant_cape_2021}, we propose a system of hybrid adversarial and local differential privacy designed for maximally private outcomes against a known set of re-identification tasks, extensible to arbitrary model architectures.

\section{Methodology}
\label{sec:methodology}

We present an overview of our empirical method for establishing the relative effect of privacy-preserving methods on the ability of an attacker to recover the value of indicator variables from intermediate representations used in a typical NLP task. For this, we rely on the Trustpilot dataset collected by \citet{Hovy}, which captures review texts and scores for various forms of consumer goods and online services, along with a range of demographic information about the review author. A short description of the dataset is in Section \ref{sec:dataset}, with a full analysis in Appendix \ref{app:dataset}.

We define as our base task sentiment prediction, that is inferring a 5-point rating from the linked review text. The text representation generated by a feature extraction component consisting of one of a set of pre-trained language models (for a list see Section \ref{sec:embeddings}) along with two densely-connected layers is given as input to the base classification model. Categorical cross-entropy is used as loss function during traning. 

In order to determine how successful an adversary may be in recovering private information from the intermediate representations within our model, we define a secondary classification task with access to our model's internal state and demographic labels for each row in our dataset (see Section \ref{sec:evaluation} for more details). For each of our private indicator variables, here gender, country, and age rank, we create a separate classification model with identical makeup to the base classifier, which will simultaneously attempt to learn the values of private indicators from the input representation.

\begin{figure}
    \centering
    \includegraphics[width=0.7\textwidth]{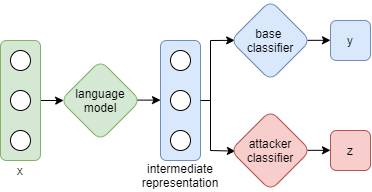}
    \caption{Indicative model architecture for base and simulated attacker tasks, where x is the input text, y is the review rating score, and z is the private attribute label.}
    \label{fig:base_model_arch}
\end{figure}

We set the model parameters according to those used in \citet{beigi_privacy_2019}, with our sentiment classifier consisting of a 200-unit hidden layer connected to a softmax output layer. To maintain constructive comparisons, we do not alter the characteristics of the classification layers during hyper-parameter tuning. A diagram of an indicative model architecture can be found in Figure \ref{fig:base_model_arch}.

\label{sec:beigi_model}
    As a baseline for comparison, we reproduce the auto-encoder-based feature extraction and classification model in \citet{beigi_privacy_2019}, under which the output of a trained encoder is substituted for the sequence embedding drawn from a language model. We have re-implemented this work based on descriptions included in the paper, deriving a network based on a gated recurrent unit (GRU) encoder and decoder setup which is trained to produce a document encoding which is then used as input to the same classification setup as in the other baseline. 

We do not compare directly to the results reported in the original research, since we are reporting results across multiple languages with a different method for determining our dataset splits, as reported in Section \ref{sec:dataset}. We compare our results only with non-privatised representations, since we wish to only establish  comparable performance in the classification part of our task set.

Once we establish the performance of our classification network, both base task and simulated attacker, for each of our test set of language models, we introduce our privacy-preserving methods. These can be broadly sorted into adversarial, differentially-private, and hybrid methods, and are described in full detail in Section \ref{sec:methodology_privacy}. Our model architecture remains static during experimentation, aside from the specific additional steps listed for each privacy method.

\subsection{Dataset}
\label{sec:dataset}
 
The Trustpilot dataset consists of user review texts along with accompanying 5-point scale scores for hundreds of thousands of users of the online consumer reviews sites collected by \citet{Hovy} in 2015. While Trustpilot operates across many countries, the researchers restricted their data collection to those countries with more than 250,000 users at the time of collection: the United Kingdom, the United States, Germany, France, and Denmark. These countries do represent significant linguistic variety, since the user is not constrained to write their review in the prevailing language of the country in which they are posting.

Demographic variables are also included in this dataset. Users can choose to add their gender, birth year, and place of residence when posting their review. The researchers also partially augment these annotations by inferring gender from the existing national distribution of first names, and by using geographical databases to convert the free text location field into a latitude/longitude pair.

For a longer description and some indicative statistics, consult Appendix \ref{app:dataset}.
 
\subsubsection{Preprocessing}

From the Trustpilot dataset, we sampled only the instances of reviews that have an attached gender and birth year annotation for the user retained. Precise location information is extremely sparse for some countries, and notably absent completely for France, and hence the country annotation is used as the target attribute for location inference. The number of total instances per country is listed in Table \ref{tab:instances}.

\begin{table}[ht]
    \footnotesize
    \centering
    \begin{tabular}{l|cc}
    \toprule
         Country & Instances \\
         \midrule
        Denmark & 467,392 \\
        France & 24,440 \\
        Germany & 26,810 \\
        United Kingdom & 146,381 \\
        United States & 43,879 \\
    \end{tabular}
    \caption{Data instances per country}
    \label{tab:instances}
\end{table}

We split the dataset by language of review text rather than by country of posting, leaving us with six languages of interest with significant representation within the dataset: English, French, German, Danish, and Norwegian. We also generate a 50,000-instance multilingual set equally composed of texts from each listed language. 

\begin{table}[ht]
    \footnotesize
    \centering
    \begin{tabular}{l|cccc}
    \toprule
        Language & Train & Test & Val & \textbf{Total} \\
        \midrule
        English & 136,226 & 38,921 & 19,460 & 194,607 \\
        French & 16,170 & 4,620 & 2,310 & 23,100 \\
        German & 20,129 & 5,751 & 2,875 & 28,755 \\
        Danish & 259,866 & 74,247 & 37,123 & 371,236 \\
        Norwegian & 51,209 & 14,630 & 7,315 & 73,154 \\
        Multi-language & 35,000 & 10,000 & 5,000 & 50,000 \\
    \end{tabular}
    \caption{Data instances per language}
    \label{tab:splits}
\end{table}

The dataset is processed so that we retain the country and gender annotations from the original dataset as our location and gender indicator variables, while birth years are used to sort the writers into three age rank bins (younger than 36, 36-45, older than 46) according to their age in 2021. We treat these features as categorical labels, converting them to an integer representation and applying one-hot encoding for use in learning. Further information on processing the data can be found in \citet{plant_cape_2021}, which describes our preliminary set of experiments using only a single set of pre-trained embeddings and English language instances from the dataset.

Splits are calculated using the adversarial Wasserstein process proposed by \citet{sogaard_we_2021}: the input texts are arranged in Wasserstein space, a random centroid is selected and the nearest neighbours form the new test set.  We use a 70/10/20 train/validation/test split structure, with the number of instances in each split shown in Table \ref{tab:splits}.

\subsection{Privacy Methods}
\label{sec:methodology_privacy}

We present here an overview of the privacy-preserving models chosen to implement as part of our suite of test scenarios.

\subsubsection{Adversarial}

\begin{figure}[ht]
    \centering
    \includegraphics[width=0.45\textwidth]{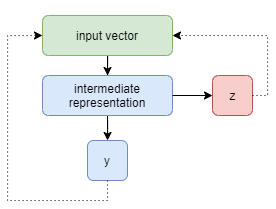}
    \caption{Adversarial training with output label $y$, and private variable label $z$. Dotted lines represent gradient updates.}
    \label{fig:adv_training}
\end{figure}

Following the 'multi-detasking' conception of \citet{coavoux_privacy-preserving_2018}, also referred to as Gradient Reversal (GR), we define the adversarial objective as the performance of our secondary classifier in recovering the label of our private variable from the intermediate representation, scored by categorical cross-entropy. We incorporate this extra information by adding a penalty to our main model when this adversary performs well, as shown in Figure \ref{fig:adv_training}. 

The loss function of a classification model parameterised by the variable $\theta_c$ can be combined with an adversarial classifier parameterised by $\theta_a$ to create an attribute-invariant model as shown in Equation \ref{def:adv_training}. 

Where $x_e$ is an example from our training set, $y$ is the target prediction label, $z$ is the label of the desired invariant feature, $\lambda$ is a regularization parameter,

\begin{equation}
\label{def:adv_training}
\mathcal{L}(x_e, y, z; \theta_c,\theta_a) = -log P(y|x_e;\theta_c) -\lambda log P(\neg z|x_e;\theta_a)
\end{equation}

We also consider an alternate formulation of the adversarial method proposed in \citet{shankar_generalizing_2018}, which they term Cross-Gradient Training (CGT). The goal of CGT is not to suppress the signal that allows the secondary classifier to recover domain information entirely, since this may remove useful information that may be germane to our network's basic task. Instead, CGT involves an attempt to augment the input with a set of perturbations that allow it to generalise better to unseen domains, learned from the gradient of loss within the secondary classifier. This notion extends to privacy in the same sense as the previous example: if we can learn a generalised version of the private information that does not directly identify the demographic label, we can achieve an acceptable level of privacy.

For each example $x$, label $y$, and private label $z$ in our sample, we carry out the following procedure: features are extracted and the private label is learned by a classification model $C_p(x)$. We then attempt to generate a new input $x'$ which would provoke a prediction from $C_p(x')$ as far as possible from $z$. The standard classification task predicting $y$ with $C_b(x')$ is also trained at the same time.

Given labelled data $x$, $y$, $z$, step size $\epsilon$, augmentation weight $\alpha$, learning rate $\eta$, and base task and private variable classifiers with parameters $\theta_b$ and $\theta_p$ and loss functions $\mathcal{L}_b$ and $\mathcal{L}_p$ respectively, we train thus,

\begin{algorithm}
\item[1] Initialise $\theta_b$ and $\theta_p$
\item[2] Sample labelled batch ($X$, $Y$, $Z$)
\item[3] $X_p := X + \epsilon \cdot \nabla_X \mathcal{L}_p(X,Z;\theta_p)$
\item[4] $X_b := X + \epsilon \cdot \nabla_X \mathcal{L}_b(X,Y;\theta_b)$
\item[5] $\theta_b \leftarrow \theta_b - \eta \nabla_{\theta_b} ((1-\alpha)\mathcal{L}_b(X, Y; \theta_b) + \alpha \mathcal{L}_b(X_b, Y; \theta_b))$
\item[6] $\theta_p \leftarrow \theta_p - \eta \nabla_{\theta_p} ((1-\alpha)\mathcal{L}_p(X, Z; \theta_p) + \alpha \mathcal{L}_p(X_p, Z; \theta_p))$
\end{algorithm}

In cases where $y$ and $z$ are highly correlated, then CGT is equivalent to traditional adversarial training. In all other cases, it will allow the model to generalise more effectively \cite{shankar_generalizing_2018}.

\subsubsection{Differential privacy}

In order to explore the potential of differential privacy, we extend the approach of \citet{lyu_differentially_2020} in using local DP to apply perturbations to the representations retrieved from our pre-trained language models, thereby reducing the certainty of a potential attacker about the true value of any recovered information, since they cannot reliably determine its difference from the other records in the dataset -- what is often referred to in the literature as indistinguishability \cite{chatzikokolakis_constructing_2015, jorgensen_conservative_2015}. 

Converting the embeddings retrieved from our language model into a DP-compliant representation requires us to inject calibrated Laplace noise into the hidden state vector obtained from the pre-trained language model as follows:

\begin{equation}
\tilde{x}_e = x_e + n
\end{equation}

where $n$ is a vector of equal length to $x_e$ containing independent and identically distributed random variables sampled from the Laplace distribution centred around 0 with a scale defined by $\frac{\Delta f}{\epsilon}$, where $\epsilon$ is the privacy budget parameter and $\Delta f$ is the sensitivity of our function. 

Since determining the sensitivity of an unbounded embedding function is practically infeasible, we constrain the range of our representation to [0,1], as recommended by \citet{shokri_privacy-preserving_2015}. In this way, the L1 norm and the sensitivity of our function summed across $n$ dimensions of $x_e$ are the same, i.e. \( \Delta f = 1 \). 

Given labelled data $x, y, z$, we follow the training process,

\begin{algorithm}
\item[1] Extract features from input sequence: $x_e = f(x)$
\item[2] Normalise representation: $x_e \leftarrow x_e - \min{x_e} / (\max{x_e} - \min{x_e}$)
\item[3] Apply perturbation: $\tilde{x}_e = x_e + Lap(\frac{\Delta f}{\epsilon})$
\item[4] Train classifiers: $\mathcal{L}(\tilde{x}_e, y, z; \theta_r, \theta_p) = -log P(y|\tilde{x}_e;\theta_r, \theta_p)$
\end{algorithm}

We also investigate the potential of metric DP (MDP) \cite{chatzikokolakis_broadening_2013} through an implementation of the work of \citet{feyisetan_privacy-_2020}. Under this regime we continue to add calibrated noise to our meaning representation---this time drawn from the multivariate normal distribution---but instead of using the noised vector as the direct input to our learning mechanism, we use the system to swap words for existing vectors that are close in the embedding space. For an input sequence $x = (w_1, w_2, \ldots, w_n)$, language model $\varphi$ with existing dictionary $D$, and privacy parameter $\epsilon$, we obtain a new sequence thus,

\begin{algorithm}
\item[1] for $i$ $\in$ ${1, ..., n}$ do
\item[2] Get embedding vector $v_i$ = $\varphi(w_i)$
\item[3] Sample noise vector $v_n = \frac{1}{(2\pi)^{m/2} \mid\Sigma\mid^{1/2}} \exp(-\frac{1}{2}(x - \mu)^T\Sigma^{-1}(x-\mu))$, where $m$ is the dimensionality of the embedding, $\Sigma$ is the covariance matrix set as identity, and the mean $\mu$ is centred at the origin
\item[4] Sample a magnitude $l$ from the Gamma distribution $l = \frac{x^{m-1}e^{-x/\phi}}{\Gamma(m)\phi^m}$, where $\phi = 1/\epsilon$
\item[5] Scale the noise vector $v_n = v_n * l$
\item[6] Perturb embedding $\hat{v_i} = v_i + v_n$ 
\item[7] Get the nearest word to perturbed vector $\hat{w_i} = argmin_{d\in D}\parallel \varphi(d) - \hat{v_i} \parallel$
\item[8] Swap $x_i$ for $\hat{w_i}$
\item[9] end
\end{algorithm}

Note that dependent on the magnitude of the noise added to the sequence and the size of the dictionary of pre-trained embeddings, the closest existing representation in embedding space to the noised word may be the original word. 

\subsubsection{Hybrid}

\begin{figure}[ht]
    \centering
    \includegraphics[width=0.9\textwidth]{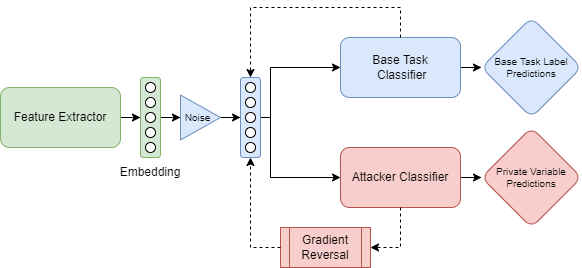}
    \caption{CAPE model diagram. Dashed lines represent gradient updates.}
    \label{fig:cape_diagram}
\end{figure}

To preserve the general privacy benefits of DP-compliant embeddings with invariance to the specific private variable identified for adversarial training, we combine both processes in a system we call Context-Aware Private Embeddings (CAPE) \cite{plant_cape_2021}. We add calibrated noise drawn from the Laplace distribution to the pre-trained sequence embedding obtained from the language model, as in \citet{lyu_differentially_2020}, as well as using the joint loss function with gradient reversal stemming from \citet{coavoux_privacy-preserving_2018}. A diagram of the combined system is pictured in Figure \ref{fig:cape_diagram}.

\subsection{Privacy Models}

We describe here the models we have chosen to apply in our experimental testing, the method of operation of which are detailed in Section \ref{sec:methodology_privacy}. 'Model' here refers to the process applied to our learning system to reduce the ability of our simulated attacker network to correctly predict the demographic attribute of interest from an intermediate representation. A listing of the models and the linked abbreviation we will use to stand in for each in reporting our results can be found in Table \ref{tab:table_key}.

\begin{table}[ht]
    \footnotesize
    \centering
    \begin{tabular}{cc}
    \toprule
        Model abbreviation & Description \\
    \midrule
        Baseline (AE) & auto-encoder-based embeddings w/ non-private classification\\
        Baseline (FX) & pre-trained embeddings w/ non-private classification\\
        CGT & Cross-gradient training\\
        GR & Gradient reversal\\
        LDP & Local differential privacy\\
        MDP & Metric differential privacy\\
        CAPE & Context-Aware Private Embeddings\\
    \end{tabular}
    \caption{Privacy model abbreviations}
    \label{tab:table_key}
\end{table}

We also include a baseline, non-private solution for comparison---the quantum of privacy protection in our system essentially being denoted by the difference in performance of our attacker on a privatised network versus the non-private. Our non-private baseline network for all languages is based on the work of \citet{guo_deep_2019}: input to our network is the pre-trained sequence embedding drawn from the language model, which is passed through two 64-unit densely connected layers, the output of which is connected to two classifier heads, one for the base task and one for the simulated attacker. Each classifier head consists of a single 200-unit dense layer connected to a softmax output layer, as shown in Figure \ref{fig:base_model_arch}.

\subsection{Language Models}
\label{sec:embeddings}

We present here the language models chosen to evaluate as a source for pre-trained text representations which may present a privacy risk. A full description of each chosen embedding source can be seen in Appendix \ref{app:lang_models}.

Language models were chosen based on the number of downloads listed on the Huggingface model repository website\footnote{\url{https://huggingface.co/models?language=en&sort=downloads}}, which we consider a reasonable proxy for popularity/impact. We rank the top five models for the English language\footnote{We disregard distilled versions since we intend to measure broad differences across architectures} and attempt to find matching models that have been developed for each of the other languages of interest. This process led us to select models based on BERT \cite{devlin_bert_2019}, RoBERTa \cite{liu_roberta_2019}, AlBERT \cite{lan_albert_2019}, and GPT-2 \cite{Radford2018GPT}. We also add GloVe \cite{pennington2014glove} and Komninos \cite{komninos_dependency_2016} embeddings for English, as representatives of previous generations of pre-trained vectors versus the more recently popular language models.

This kind of parity was more difficult to achieve for multilingual models, given that successful monolingual models often do not transfer to this domain, and hence we include the 'paraphrase-xlm-r-multilingual-v1' \cite{reimers_making_2020} and 'distiluse-base-multilingual-cased-v2' \cite{yang_multilingual_2020} models.

\section{Experimental Setup}

In this section, we lay out the conditions for our empirical work, including the setup of each class of experiment, the quantification of relative degrees of private information leakage, and the evaluation framework in use.

In Section \ref{exp:baseline}, we carry out a comparison of our baseline classification method for private attributes to the auto-encoder-based method proposed by \citet{beigi_privacy_2019}, as described in Section \ref{sec:beigi_model}. We use this baseline method in successive experiments to establish a non-private benchmark for private information leakage that we can compare to each of our privatising methods to quantify their impact.

In Section \ref{exp:privacy}, we report the performance of our attacker classifier when we apply our privacy-preserving models to each language dataset individually, as well as the multilingual set. Gradient reversal (GR) and cross-gradient training (CGT) methods are applied as previously described: for GR, we simply invert the sign of the gradient during backpropagation. 
CGT requires us to watch the gradient of the attacker classifier during training to perturb the input to both classifiers. We set the step size of perturbation during training to $5$ and the $\alpha$ parameter, which controls the ratio of the contribution of the perturbed input versus the original input to the total loss, to $0.5$.

In terms of our locally differentially private models, both LDP and CAPE, we add calibrated noise from the Laplace distribution to our sequence embedding between the output of our language model and the first dense layer of our network. The $\epsilon$ parameter controlling the size of our privacy budget is kept static at $0.1$ throughout this stage of testing. In order to test metric differential privacy (MDP), we instead pre-compute all sequence swaps across our dataset before training or inference, since exhaustively searching the embedding space for every word vector during online training proved highly computationally expensive. In this case, the $\epsilon$ privacy parameter is set to a value of $20$ during testing. Base and adversary classifiers are trained and tested with noised sequences obtained in these ways in the same fashion as the other experimental models.

Embodied in each of our privatisation strategies is a decision about the level of performance degradation we will accept to decrease privacy risk. In order to quantify the ratio between enhanced attribute privacy and decreased predictive power, we must measure the utility of our network. The effect of each privacy-preserving method as represented by performance in our basic task across our main set of experiments are reported in Section \ref{exp:utility}.

\subsection{Evaluation Criteria}
\label{sec:evaluation}

We consider the primary metric of success in our attacker task as the ability to reliably distinguish the gender, age, or location of the individual from the text representation. We establish the rate of successful predictions of our demographic attribute as a measure of the privacy at risk, as in \citet{jaiswal_privacy_2020, zhao_gender_2019} -- any change in this from our baseline to private methods will indicate this. Since we are dealing with an imbalanced dataset across multiple axes of comparison, we report both accuracy and F1-scores, emphasising the latter as a more reliable measure. 

We provide summary statistics for each experiment using the two-way ANOVA method, reporting the variance in F1 results ($\sigma^2$) across the private information type categorical variable row-wise, and the summary sum of squares (SS), degrees of freedom(df), F-value (F), F critical value (F-crit) and p-value (P) scores both row- and across the embedding type column-wise, and across the whole set of results. F-crit here is calculated based on a 0.05 p-value significance threshold.

Since we rely on empirical demonstrations of our privacy outcomes, we also consider utility through the same lens. Performance in the downstream task, prediction of the review score class from the text as laid out in Section \ref{sec:methodology}, forms our primary metric in comparison to the baseline model. We report F1-score as a robust mechanism for use across our test scenarios. 

\section{Results}
\label{sec:results}

In this section we will discuss in detail the results of each experiment, as well as discuss some initial findings from our empirical data.

\subsection{Non-private baselines}
\label{exp:baseline}

We present here the results from a comparison of our basic method to the auto-encoder-based method of \cite{beigi_privacy_2019}. It should be noted that we have not attempted to implement the privacy-enhancing elements of the work, and therefore these results establish only a comparable level of performance for both the basic sentiment analysis task and the private information re-identification task for our system with a recent published baseline. Results for the base task, as well as each private demographic variable attacker network, can be seen in Table \ref{tab:baseline}. Accuracy and F1-score are presented for each embedding class, with the best-performing instance, e.g. the classifier most able to predict the target variable, highlighted.

We apply these methods to only the English-language instances from our dataset, to more accurately represent a comparison with the original work, which included only English results from the Trustpilot set. We do however present our language-model-based approach as a non-private baseline comparison in all further experiments, regardless of language.

\begin{table}[ht]
    \footnotesize
    \centering
    \begin{tabular}{lccccccccc}
    \toprule
    & & \multicolumn{6}{c}{Private Variable}\\
    & \multicolumn{2}{c}{Base Task} & \multicolumn{2}{c}{Gender} & \multicolumn{2}{c}{Location} & \multicolumn{2}{c}{Age} & $\sigma^2$\\
    \cmidrule(lr){2-3}\cmidrule(lr){4-5}\cmidrule(lr){6-7}\cmidrule(lr){8-9}
     Model & acc. & f1 & acc. & f1 & acc. & f1 & acc. & f1\\
    \midrule
    \textbf{Baseline (FX)} &  \\
    Gl & 0.646 & 0.624 & 0.749 & 0.674 & 0.732 & 0.661 & 0.752 & 0.674 & 0.021 \\
    Ko & 0.631 & 0.598 & 0.749 & 0.668 & 0.713 & 0.630 & 0.757 & 0.680 & 0.024\\
    Be & 0.680 & 0.670 & 0.788 & 0.737 & 0.794 & 0.749 & 0.794 & 0.746 & 0.027 \\
    MB & 0.662 & 0.654 & 0.763 & 0.712 & 0.749 & 0.696 & 0.752 & 0.691 & 0.021 \\
    Ro & 0.690 & 0.681 & \textbf{0.802} & 0.765 & \textbf{0.811} & \textbf{0.771} & \textbf{0.811} & \textbf{0.773} & 0.026 \\
    Al & \textbf{0.691} & \textbf{0.685} & 0.795 & \textbf{0.779} & 0.806 & 0.765 & 0.801 & 0.754 & 0.029 \\
    GPT & 0.689 & 0.681 & 0.792 & 0.758 & 0.789 & 0.730 & 0.756 & 0.753 & 0.024 \\
    \midrule
    \textbf{Baseline (AE)} & 0.599 & 0.449 & 0.537 & 0.375 & 0.755 & 0.649 & 0.599 & 0.449 & 0.034 \\
    \end{tabular}
    \caption{Non-private task baseline comparison. Gl: GloVe, Ko: Komninos, Be: BERT, MB: Multilingual BERT, Ro: RoBERTa, Al: ALBERT, GPT: GPT-2.}
    \label{tab:baseline}
\end{table}

\begin{table}[ht]
\footnotesize
    \centering
    \begin{tabular}{lccccc}
    \toprule
         & SS & df & F & P & F-crit \\
        \midrule
        Row & 0.069 & 7 & 0.783 & 0.613 & 2.764 \\
        Column & 0.236 & 2 & 9.318 & 0.003 & 3.739 \\ 
        Total & 0.482 & 23 \\
    \end{tabular}
    \caption{Non-private baseline results summary statistics.}
    \label{tab:baseline_summary}
\end{table}

Several interesting trends can be observed immediately from inspection of the results. First, using pre-trained embeddings from popular LLMs without significant fine-tuning immediately outperforms an auto-encoder-based approach trained only on instances from our dataset. This finding is perhaps unsurprising since the training corpus and depth of the model is far less extensive. These approaches may however be eminently compatible \citep{lewis_bart_2020, gordon_long_2020, li_improving_2020}. 

It is important to note here that we do not believe it is judicious to compare directly our results to the results of \citet{beigi_privacy_2019}, as discussed in Section \ref{sec:beigi_model}. Also, it is important to bear in mind that we applied this system to a completely separate dataset split than that used in the original paper, which in our case was generated through the application of an adversarial selection process, as detailed in Section \ref{sec:dataset}. Indeed, under the reproducibility schema advocated in \cite{belz_quantifying_2021}, we would not have access to the object conditions, but rather a subset measurement method and procedure conditions.

Secondly, it appears that pre-trained embeddings that lead to higher performance on the basic sentiment analysis task also enable better performance for the attacker, as shown by the results for Roberta and Albert in Table \ref{tab:baseline}. We posit that by becoming more efficient at encoding contextual semantic information through the co-occurrence of terms in their training sets, these models have also become more vulnerable to private data leakage through also encoding elements of authorial style \cite{emmeryAdversarialStylometryWild2021}. This effect is not negligible, with the largest relative change in F1-score amounting to $\sim16\%$. 

\subsection{Privacy-preserving models}
\label{exp:privacy}

In this section we present the results for each of our previously cited privacy-preserving processes of interest, as detailed in Section \ref{sec:methodology_privacy}, applied across the various languages in our Trustpilot dataset. We investigate here not only each language corpus separately, but a combined multilingual set drawn from each. For each privacy-preserving model, the set of attacker tasks aiming to predict the gender, location, and age variable are applied separately to each set of pre-trained represetations, the score for which is presented as a row in our results tables (\ref{tab:en_single_var}, \ref{tab:fr_single_var}, \ref{tab:de_single_var}, \ref{tab:da_single_var}, \ref{tab:no_single_var}, \ref{tab:all_single_var}).

Results for private variable classifiers are presented as F1-scores. Results in the table are presented at a precision of 4 digits. Best performance---that is, showing the poorest record of predicting the target variable---is picked out in bold.

\begin{table}[ht]
    \footnotesize
    \centering
    \begin{tabular}{l|cccc}
    \toprule
    Model & Gender & Location & Age & $ \sigma^2$ \\
    \midrule
    \textbf{Baseline} &  \\
    Gl & 0.674 & 0.661 & 0.674 & 0.021 \\
    Ko & 0.668 & 0.630 & 0.680 & 0.024 \\
    Be & 0.737 & 0.749 & 0.746 & 0.027 \\
    MB & 0.712 & 0.696 & 0.691 & 0.021 \\
    Ro & 0.765 & 0.771 & 0.773 & 0.026\\
    Al & 0.779 & 0.765 & 0.754 & 0.029\\
    GPT & 0.758 & 0.730 & 0.753 & 0.024\\
    \midrule
    \textbf{CGT} &\\
    Gl & 0.375 & 0.649 & 0.449 & 0.020\\
    Ko & 0.375 & 0.649 & 0.449 & 0.020\\
    Be & 0.375 & 0.649 & 0.449 & 0.020\\
    MB & 0.375 & 0.649 & 0.449 & 0.020\\
    Ro & 0.375 & 0.649 & 0.449 & 0.020\\
    Al & 0.375 & 0.649 & 0.449 & 0.020\\
    GPT & 0.375 & 0.649 & 0.449 & 0.020\\
    \midrule
    \textbf{GR} &\\
    Gl & 0.375 & 0.649 & 0.449 & 0.020\\
    Ko & 0.378 & 0.649 & 0.449 & 0.020\\
    Be & 0.375 & 0.649 & 0.448 & 0.020\\
    MB & 0.377 & 0.649 & 0.448 & 0.020\\
    Ro & 0.375 & 0.649 & 0.449 & 0.020\\
    Al & 0.374 & 0.649 & 0.449 & 0.020\\
    GPT & 0.376 & 0.648 & 0.448 & 0.020\\
    \end{tabular}
    \quad
    \begin{tabular}{l|cccc}
    \toprule
    Model & Gender & Location & Age & $\sigma^2$ \\
    \midrule
    \textbf{LDP}\\
    Gl & - & - & -\\
    Ko & - & - & -\\
    Be & 0.375 & \textbf{0.649} & 0.449 & 0.020\\
    MB & 0.375 & 0.709 & 0.454 & 0.030\\
    Ro & 0.375 & 0.649 & 0.449 & 0.020\\
    Al & 0.646 & 0.819 & 0.457 & 0.033\\
    GPT & 0.561 & 0.708 & 0.449 & 0.017\\
    \midrule
    \textbf{MDP}\\
    Gl & 0.572 & 0.719 & 0.456 & 0.017\\
    Ko & 0.572 & 0.703 & 0.459 & 0.015\\
    Be & 0.653 & 0.798 & 0.498 & 0.023\\
    MB & 0.635 & 0.738 & 0.472 & 0.018\\
    Ro & 0.664 & 0.809 & 0.501 & 0.024\\
    Al & 0.645 & 0.790 & 0.479 & 0.024\\
    GPT & 0.646 & 0.768 & 0.491 & 0.019\\
    \midrule
    \textbf{CAPE} & \\
    Gl & - & - & -\\
    Ko & - & - & -\\
    Be & 0.507 & 0.649 & 0.339 & 0.024\\
    MB & 0.378 & 0.649 & \textbf{0.278} & 0.082\\
    Ro & \textbf{0.350} & 0.649 & 0.449 & 0.023\\
    Al & 0.426 & 0.649 & 0.424 & 0.017\\
    GPT & 0.375 & 0.649 & 0.449 & 0.020\\
    \end{tabular}
    \caption{English results. Gl: GloVe, Ko: Komninos, Be: BERT, MB: Multilingual BERT, Ro: RoBERTa, Al: ALBERT, GPT: GPT-2}
    \label{tab:en_single_var}
\end{table}

\begin{table}[ht]
\footnotesize
    \centering
    \begin{tabular}{lccccc}
    \toprule
         & SS & df & F & P & F-crit \\
        \midrule
        Row & 4.024 & 41 & 18.674 & 1.334$e^{-27}$  & 1.537 \\
        Column & 1.331 & 2 & 126.587 & 8.517$e^{-26}$ & 3.108 \\ 
        Total & 5.786 & 125 \\
    \end{tabular}
    \caption{English results summary statistics.}
    \label{tab:english_summary}
\end{table}

We can derive several interesting observations from our English-language results, displayed in Table \ref{tab:en_single_var}. Firstly, it is clear that the application of both broad categories of privacy-preserving mechanism---differential privacy noise and adversarial training---produce a marked drop in the information leaked to our simulated attacker network. In fact, in the best case, that of RoBERTa embeddings passed through the CAPE model, the F1-score of our gender attacker experiences a relative drop of $\approx 45\%$ over baseline. 

This drop is variable across the range of embeddings tested, but several tendencies can be observed in action:

\begin{itemize}
    \item \textbf{Models with more parameters show larger mean reductions in attacker performance.} We note when comparing RoBERTa, ALBERT, BERT, and GPT-2, that the results for the change in performance of gender re-identification against the baseline appear to be proportional to the size of the model in terms of number of parameters. ALBERT shows an $\approx 25\%$ average drop in performance with the smallest set of 11 million parameters, BERT shows an $\approx 30\%$ drop with 109 million, GPT-2 shows a reduction of $\approx 32\%$ with 117 million, while RoBERTa shows the largest relative drop of $\approx 39\%$ with 125 million parameters.
    \item \textbf{Adversarial gradient-based methods produce remarkably homogeneous results.} The variance between results for different sets of embeddings for our purely adversarial methods (gradient reversal and cross-gradient training), are much lower than that of our other privacy-preserving methods. Indeed, at the level of precision displayed in Table \ref{tab:en_single_var}, the vast majority of scores are identical. We theorise that this effect is due to a decrease in variance across the embeddings caused by the suppression of contextual stylometric cues that also impacts the existing contextual semantic relationships between token vectors. We explore the impact of this on downstream performance in Section \ref{exp:utility}, as well as investigating the statistical deviation of the embedding vector in Section \ref{sec:noise}.
    \item \textbf{Adding adversarial training objectives decreases information leakage in a DP-compliant scenario.} Comparing the mean performance reduction over baseline for all private demographic variables across our differentially-private systems---local differential privacy, metric differential privacy, and CAPE--we find that CAPE provides an $\approx 30\%$ reduction, while LDP and MDP show smaller drops of $\approx 19\%$ and $\approx 6\%$ respectively. It is worth pointing out that under some circumstances, represented here by the attempt to classify the location variable, CAPE displays the same previously mentioned reduction in variance as other adversarial methods.
    \item \textbf{Age re-identification performance in mBERT is low.} We note that in the case of our baseline and CAPE models, embeddings extracted from mBERT show anomalously low age identification results. One potential explanation for this outcome is the vocabulary peculiarities of mBERT; while the model vocabulary is much larger than BERT (119,000 vs. 30,000 entries), minority languages are oversampled in relation to English \cite{abdaoui_load_2020}, which could indicate that archaisms or neologisms that may indicate age are being left out.
\end{itemize}

We note here that for the non-contextual pre-trained embeddings tested (GloVe and Komninos vectors), downstream task performance was so poor once any element of differential noise was added that no useful metric could be extracted to quantify the other impacts of our privacy-preserving systems. We suggest that this indicates our privacy budget objective was set at an level inappropriate to the use of these lower-dimensional vectors, and that further analysis is required to determine a suitable standard for producing effective intermediate representations in these settings. For further discussion of the impact of privacy budgeting please consult Section \ref{sec:noise}.

\begin{table}[ht]
    \footnotesize
    \centering
    \begin{tabular}{l|cccc}
    \toprule
    Model & Gender & Location & Age & $\sigma^2$ \\
    \midrule
    \textbf{Baseline} &  \\
    FB & 0.680 & 0.985 & 0.414 & 0.082 \\
    MB & 0.658 & 0.985 & 0.405 & 0.085\\
    CB & 0.724 & 0.985 & 0.447 & 0.072\\
    GPT & 0.645 & 0.985 & 0.376 & 0.093\\
    \midrule
    \textbf{CGT} &\\
    FB & 0.404 & 0.985 & 0.419 & 0.110\\
    MB & 0.404 & 0.985 & 0.419 & 0.110\\
    CB & 0.404 & 0.985 & 0.419 & 0.110\\
    GPT & 0.404 & 0.985 & 0.419 & 0.110\\
    \midrule
    \textbf{GR} &\\
    FB & 0.410 & 0.985 & 0.414 & 0.109\\
    MB & 0.404 & 0.985 & 0.419 & 0.110\\
    CB & 0.404 & 0.985 & 0.419 & 0.110\\
    GPT & 0.405 & \textbf{0.984} & 0.405 & 0.112\\
    \end{tabular}
    \quad
    \begin{tabular}{l|cccc}
    \toprule
    Model & Gender & Location & Age & $\sigma^2$ \\
    \midrule
    \textbf{LDP}\\
    FB & 0.405 & 0.985 & 0.419 & 0.110\\
    MB & 0.404 & 0.985 & 0.419 & 0.110\\
    CB & 0.404 & 0.985 & 0.419 & 0.110\\
    GPT & 0.636 & 0.985 & 0.419 & 0.082\\
    \midrule
    \textbf{MDP}\\
    FB & 0.571 & 0.985 & 0.417 & 0.086\\
    MB & 0.624 & 0.985 & 0.419 & 0.082\\
    CB & 0.712 & 0.985 & 0.443 & 0.073\\
    GPT & 0.659 & 0.985 & 0.414 & 0.082\\
    \midrule
    \textbf{CAPE} & \\
    FB & 0.499 & 0.985 & \textbf{0.383} & 0.102\\
    MB & 0.404 & 0.985 & 0.404 & 0.113\\
    CB & \textbf{0.403} & 0.985 & 0.414 & 0.111\\
    GPT & 0.435 & 0.985 & 0.419 & 0.104\\
    \end{tabular}
    \caption{French results. GPT: GPT-2, MB, Multilingual BERT, CB: CamemBERT, FB: FlauBERT}
    \label{tab:fr_single_var}
\end{table}

\begin{table}[ht]
\footnotesize
    \centering
    \begin{tabular}{lccccc}
    \toprule
         & SS & df & F & P & F-crit \\
        \midrule
        Row & 0.125 & 23 & 1.047 & 0.434  & 1.767 \\
        Column & 4.514 & 2 & 435.333 & 1.3$e^{-30}$ & 3.200 \\ 
        Total & 4.878 & 71 \\
    \end{tabular}
    \caption{French results summary statistics.}
    \label{tab:fr_summary}
\end{table}

\begin{table}[ht]
    \footnotesize
    \centering
    \begin{tabular}{l|cccc}
    \toprule
    Model & Gender & Location & Age & $\sigma^2$ \\
    \midrule
    \textbf{Baseline} &  \\
    Be & 0.683 & 0.985 & 0.391 & 0.088\\
    MB & 0.614 & 0.985 & 0.413 & 0.084\\
    Ro & 0.612 & 0.985 & 0.435 & 0.079\\
    GPT & 0.677 & 0.985 & 0.449 & 0.072\\
    \midrule
    \textbf{CGT} &\\
    Be & 0.528 & 0.985 & 0.328 & 0.113\\
    MB & 0.528 & 0.985 & 0.328 & 0.113\\
    Ro & 0.528 & 0.985 & 0.328 & 0.113\\
    GPT & 0.528 & 0.985 & 0.328 & 0.113\\
    \midrule
    \textbf{GR} &\\
    Be & 0.528 & 0.985 & \textbf{0.146} & 0.177\\
    MB & 0.529 & 0.985 & 0.249 & 0.138\\
    Ro & 0.544 & 0.984 & 0.317 & 0.115\\
    GPT & 0.528 & 0.985 & 0.328 & 0.113\\
    \end{tabular}
    \quad
    \begin{tabular}{l|cccc}
    \toprule
    Model & Gender & Location & Age & $\sigma^2$ \\
    \midrule
    \textbf{LDP}\\
    Be & 0.528 & 0.985 & 0.328 & 0.113\\
    MB & 0.528 & 0.985 & 0.328 & 0.113\\
    Ro & 0.528 & 0.985 & 0.328 & 0.113\\
    GPT & 0.528 & 0.985 & 0.328 & 0.113\\
    \midrule
    \textbf{MDP}\\
    Be & 0.673 & 0.985 & 0.401 & 0.086\\
    MB & 0.632 & 0.985 & 0.421 & 0.081\\
    Ro & 0.640 & 0.985 & 0.440 & 0.076\\
    GPT & 0.680 & 0.985 & 0.424 & 0.079\\
    \midrule
    \textbf{CAPE} & \\
    Be & 0.528 & 0.985 & 0.328 & 0.114\\
    MB & 0.528 & 0.985 & 0.328 & 0.113\\
    Ro & \textbf{0.527} & 0.985 & 0.328 & 0.114\\
    GPT & 0.528 & 0.985 & 0.198 & 0.156\\
    \end{tabular}
    \caption{German results. Be: BERT, MB: Multilingual BERT, Ro: RoBERTa, GPT: GPT-2}
    \label{tab:de_single_var}
\end{table}

\begin{table}[ht]
\footnotesize
    \centering
    \begin{tabular}{lccccc}
    \toprule
         & SS & df & F & P & F-crit \\
        \midrule
        Row & 0.121 & 23 & 2.717 & 0.002 & 1.767 \\
        Column & 5.098 & 2 & 1314.380 & 2.61$e^{-41}$ & 3.200 \\ 
        Total & 5.308 & 71 \\
    \end{tabular}
    \caption{German results summary statistics.}
    \label{tab:de_summary}
\end{table}

We consider here the results for French and German language texts, presented in Tables \ref{tab:fr_single_var} and \ref{tab:de_single_var}. We claim that the language models used to generate embeddings for these languages represent a fair comparison to the English language models used to generate the previous set of results, since the training data corpus for each is comparable in size and scope. For instance, the RoBERTa-derived French model CamemBERT \cite{martin_camembert_2020} was trained using 138GB of French documents from the multi-lingual OSCAR dataset \cite{OrtizSuarezSagotRomary2019}, which compares favourably to the mixed 160GB of documents in the original model training corpus \cite{liu_roberta_2019}. This holds true also for FlauBERT \cite{le_flaubert_2020} and German BERT \cite{chan_germans_2020} as compared to the original BERT (71GB and 12GB vs. 13GB). On this basis, we believe that the set of models used in these tests can be considered as relatively even in terms of language resources, and hence we believe that any observed deviation of results can be attributed to changes in the language structure and textual content, rather than the effect of sparse training/fine-tuning text availability. 

We note the following interesting findings from these experiments:

\begin{itemize}
    \item \textbf{Re-identification of gender is higher on average than English results.} The mean average results across all privacy-preserving technologies is higher for both French ($\approx 4\%$ relative) and German ($\approx 20\%$ relative) when compared to English. French and German exhibit more gendered language features than English \cite{hord_bucking_2016, kokovidis_gendered_2015, fiedler_representation_2007}.
    \item \textbf{Performance of age identification across all models is much lower.} Our results show an absolute drop in age re-identification performance of $\approx 4\%$ for French and $\approx 11\%$ for German. We note that this may be due to the increased prevalence of fake data for this attribute linked to reduced willingness to share accurate personal information in these nations, as discussed in Section \ref{sec:privacy_risk}. 
    \item \textbf{Location in terms of country ceases to be a meaningful target variable.} While testing with English-language instances we could expect significant proportions of source texts to be tagged with countries other than the UK, since Trustpilot also operates in numerous other nations where English is the primary language, including the United States. We can no longer make that statement confidently for French or German, where we would expect the vast majority to be sourced from the country in question. In that case, high accuracy re-identification by classification is achievable by brute force guessing. In these cases, identifying the country of origin does not represent an increased privacy threat - the text itself does that reliably. Regional and locality identification would still present a threat, although one not addressed in this experiment. 
\end{itemize}

Overall, we continue to note the reduced performance of attacker networks in the CAPE setting over other differentially-private models, which continues to demonstrate the effectiveness of adversarial training at reducing information leakage.

\begin{table}[ht]
    \footnotesize
    \centering
    \begin{tabular}{l|cccc}
    \toprule
    Model & Gender & Location & Age & $\sigma^2$ \\
    \midrule
    \textbf{Baseline} &  \\
    Be & 0.689 & 0.996 & 0.392 & 0.091 \\
    MB & 0.622 & 0.996 & 0.334 & 0.110 \\
    Ro & 0.584 & 0.996 & 0.300 & 0.122 \\
    \midrule
    \textbf{CGT} &\\
    Be & 0.430 & 0.996 & 0.272 & 0.145 \\
    MB & 0.430 & 0.996 & 0.272 & 0.145 \\
    Ro & 0.430 & 0.996 & 0.272 & 0.145 \\
    \midrule
    \textbf{GR} &\\
    Be & 0.430 & 0.996 & \textbf{0.010} & 0.249 \\
    MB & 0.430 & 0.996 & 0.262 & 0.148 \\
    Ro & 0.430 & 0.996 & 0.098 & 0.206 \\
    \end{tabular}
    \quad
    \begin{tabular}{l|cccc}
    \toprule
    Model & Gender & Location & Age & $\sigma^2$ \\
    \midrule
    \textbf{LDP}\\
    Be & 0.608 & 0.996 & 0.341 & 0.108 \\
    MB & 0.430 & 0.996 & 0.225 & 0.159 \\
    Ro & 0.430 & 0.996 & 0.257 & 0.149 \\
    \midrule
    \textbf{MDP}\\
    Be & 0.692 & 0.996 & 0.415 & 0.084 \\
    MB & 0.649 & 0.996 & 0.337 & 0.109 \\
    Ro & 0.535 & 0.996 & 0.260 & 0.138 \\
    \midrule
    \textbf{CAPE} & \\
    Be & 0.430 & 0.996 & 0.121 & 0.197 \\
    MB & \textbf{0.430} & 0.996 & 0.197 & 0.169 \\
    Ro & 0.430 & \textbf{0.995} & 0.272 & 0.145 \\
    \end{tabular}
    \caption{Danish results. Be: BERT, MB: Multilingual BERT, Ro: RoBERTa}
    \label{tab:da_single_var}
\end{table}

\begin{table}[ht]
\footnotesize
    \centering
    \begin{tabular}{lccccc}
    \toprule
         & SS & df & F & P & F-crit \\
        \midrule
        Row & 0.208 & 17 & 2.688 & 0.007 & 1.933 \\
        Column & 5.082 & 2 & 559.094 & 9.755$e^{-27}$ & 3.276 \\ 
        Total & 5.444 & 53 \\
    \end{tabular}
    \caption{Danish results summary statistics.}
    \label{tab:da_summary}
\end{table}

\begin{table}[ht]
    \footnotesize
    \centering
    \begin{tabular}{l|cccc}
    \toprule
    Model & Gender & Location & Age & $\sigma^2$  \\
    \midrule
    \textbf{Baseline} &  \\
    Be & 0.590 & 0.995 & 0.308 & 0.119 \\
    MB & 0.589 & 0.995 & 0.310 & 0.118 \\
    Ro & 0.620 & 0.995 & 0.301 & 0.120 \\
    \midrule
    \textbf{CGT} &\\
    Be & 0.407 & 0.995 & 0.254 & 0.153 \\
    MB & 0.407 & 0.995 & 0.254 & 0.153 \\
    Ro & 0.407 & 0.995 & 0.254 & 0.153 \\
    \midrule
    \textbf{GR} &\\
    Be & 0.407 & 0.995 & 0.230 & 0.160 \\
    MB & 0.407 & \textbf{0.994} & 0.152 & 0.187 \\
    Ro & 0.407 & 0.995 & \textbf{0.087} & 0.212 \\
    \end{tabular}
    \quad
    \begin{tabular}{l|cccc}
    \toprule
    Model & Gender & Location & Age & $\sigma^2$  \\
    \midrule
    \textbf{LDP}\\
    Be & 0.407 & 0.995 & 0.249 & 0.154 \\
    MB & 0.407 & 0.995 & 0.206 & 0.168 \\
    Ro & 0.407 & 0.995 & 0.212 & 0.166 \\
    \midrule
    \textbf{MDP}\\
    Be & 0.606 & 0.995 & 0.285 & 0.126 \\
    MB & 0.571 & 0.995 & 0.279 & 0.129 \\
    Ro & 0.612 & 0.995 & 0.276 & 0.129 \\
    \midrule
    \textbf{CAPE} & \\
    Be & 0.326 & 0.995 & 0.252 & 0.167 \\
    MB & 0.453 & 0.995 & 0.203 & 0.164 \\
    Ro & \textbf{0.309} & 0.995 & 0.156 & 0.199 \\
    \end{tabular}
    \caption{Norwegian results. Be: BERT, MB: Multilingual BERT, Ro: RoBERTa}
    \label{tab:no_single_var}
\end{table}

\begin{table}[ht]
\footnotesize
    \centering
    \begin{tabular}{lccccc}
    \toprule
         & SS & df & F & P & F-crit \\
        \midrule
        Row & 0.127 & 17 & 2.211 & 0.024 & 1.933 \\
        Column & 5.443 & 2 & 803.777 & 2.376$e^{-29}$ & 3.276 \\ 
        Total & 5.685 & 53 \\
    \end{tabular}
    \caption{Norwegian results summary statistics.}
    \label{tab:no_summary}
\end{table}

Turning to the results for Danish and Norwegian dataset instances, as displayed in Tables \ref{tab:da_single_var} and \ref{tab:no_single_var}, we claim that these can be understood as representing a typical scenario for medium-resource languages \cite{ortiz_suarez_monolingual_2020, iliev_expanding_2012}. In these cases, the language model has been pre-trained in the same way as the original research in a high-resource setting, but with a substantially reduced text corpus. The Danish BERT model used in these experiments for instance is trained on around 1.6 billion words of text, while the original BERT model was trained with around 3.3 billion words. This was also the case for the NorBERT model used to produce Norwegian embeddings, which was trained with around 2 billion words. 

In this way, we can expect the model to be less sensitive to context variation and less able to represent complex text sequences, since we would intuitively expect general task performance to scale with corpus size and variability \cite{sasano_effect_2009, rose_effects_1997}.

\begin{itemize}
    \item \textbf{Location prediction continues to be a trivial task.} As can be seen in the results for both languages, prediction scores continue to trend towards perfect performance, since a negligible number of the dataset instances originate from nations other than the country of language origin. Mean performance across all models for both languages (0.995) reaches a marginally higher rate than the French and German examples (0.985), which indicates that the instances are more highly localised than the major western European languages.
    \item \textbf{Gender re-identification performance is lower than higher-resource settings.} Comparing the performance across all embeddings of our gender prediction task, we can see that both Danish and Norwegian show comparatively lower scores than any previously analysed language set. In terms of mean F1-score, Danish shows an $\approx 3\%$ absolute drop in performance, while Norwegian shows a larger $\approx 6\%$ absolute reduction. This lower performance ratio is mirrored in the results once privacy-preservation is applied. This result could be linked to the lower sensitivity of the model due to limited training texts, as previously mentioned, although we also contend that ongoing changes in the contextual and grammatical markers of gender in online language use for these languages could also be a factor \cite{lohndal_grammatical_2021, cornips_comparative_2017, rodina_grammatical_2015}.
    \item \textbf{Accuracy of age classification is lower than all other experiments.} We note that the performance of our re-identification network in the majority of settings is lower by an average of $>10\%$ for both lower-resource languages than the other language settings studied.\\ This may reflect the a higher pitch of public reticence to sharing age data as has been mentioned previously, or may reflect a lack of variation in the training corpora; it could be, for instance, that the Danish and Norwegian sections of Commmon Crawl and Wikipedia do not contain as much stylistic variation between authors of differing age ranges as the larger English or French sections.
\end{itemize}

\begin{table}[ht]
    \footnotesize
    \centering
    \begin{tabular}{l|cccc}
    \toprule
    Model & Gender & Location & Age & $\sigma^2$ \\
    \midrule
    \textbf{Baseline} &  \\
    MB & 0.589 & 0.598 & 0.589 & 2.665$e^{-5}$\\
    USE & 0.617 & 0.612 & 0.608 & 1.847$e^{-5}$\\
    XLM & 0.640 & 0.638 & 0.643 & 5.078$e^{-6}$\\
    \midrule
    \textbf{CGT} &\\
    MB & 0.412 & 0.859 & 0.336 & 0.080 \\
    USE & 0.530 & 0.768 & 0.336 & 0.047 \\
    XLM & 0.574 & 0.749 & 0.336 & 0.043 \\
    \midrule
    \textbf{GR} &\\
    MB & 0.412 & 0.576 & 0.336 & 0.015 \\
    USE & 0.412 & \textbf{0.439} & 0.335 & 0.003 \\
    XLM & 0.412 & 0.440 & 0.320 & 0.004 \\
    \end{tabular}
    \quad
    \begin{tabular}{l|cccc}
    \toprule
    Model & Gender & Location & Age & $\sigma^2$ \\
    \midrule
    \textbf{LDP}\\
    MB & 0.412 & 0.860 & 0.336 & 0.080 \\
    USE & 0.412 & 0.439 & 0.336 & 0.003 \\
    XLM & 0.420 & 0.718 & 0.325 & 0.042 \\
    \midrule
    \\
    \\
    \\
    \\
    \midrule
    \textbf{CAPE} & \\
    MB & 0.412 & 0.842 & 0.336 & 0.075 \\
    USE & \textbf{0.412} & 0.439 & 0.316 & 0.004 \\
    XLM & 0.460 & 0.439 & \textbf{0.266} & 0.011 \\
    \end{tabular}
    \caption{Multi-language results. MB: Multilingual BERT, XLM: Multilingual RoBERTa, USE: Multilingual Universal Sentence Encoder}
    \label{tab:all_single_var}
\end{table}

\begin{table}[ht]
\footnotesize
    \centering
    \begin{tabular}{lccccc}
    \toprule
         & SS & df & F & P & F-crit \\
        \midrule
        Row & 0.333 & 14 & 1.863 & 0.078 & 2.064 \\
        Column & 0.456 & 2 & 17.872 & 9.957$e^{-6}$ & 3.340 \\ 
        Total & 1.147 & 44 \\
    \end{tabular}
    \caption{Multi-language results summary statistics.}
    \label{tab:all_summary}
\end{table}

Finally, we present the results for privacy models trained using a multi-lingual corpus. We can expect this model to show lower baseline performance and sensitivity overall than models trained using a large corpus of documents in the individual languages \cite{pires_how_2019, rust_how_2021}. However, we do establish by examining the baseline results that cross-lingual demographic prediction is viable given a multilingual embedding, indicating that such models are vulnerable to the same information leakage as monolingual language models.

In addition we note that:

\begin{itemize}
    \item \textbf{Location prediction is viable once more.} Prediction at the national level across the set of languages in our dataset proves meaningful once more, given the heterogeneous set of localities in the test set. We note that certain applications of privacy-preserving systems here enhance the ability of our attacker network in specific settings---for instance when applying differential privacy across mBERT embeddings, or using cross-gradient training. We suggest that this effect may be due to the additional noise pushing the distinct clusters of single language embeddings further apart in a way which may make them easier for our attacker network to classify. This possibility will be examined in more detail in Section \ref{sec:noise}.
    \item \textbf{Gender results trend lower than monolingual models.} Results for gender re-identification more closely resemble the Danish and Norwegian scenarios than the relatively higher-resource languages. We suggest that this may point to a commonality: that models with a larger and more diverse training corpus in a target language are better able to represent the gendered nature of textual sequences, and hence are more liable to leak information. 
    \item \textbf{Age prediction results are consistent with higher-resource settings.} Age re-identification scores show a similar distribution to the relatively higher-resource settings of French and German, rather than the very poor performance seen in Danish and Norwegian. 
\end{itemize}

We were unable to provide results for Metric Differential Privacy in this setting due to the technical complexity involved in projecting multiple language training sets into a common embedding space. We expect to address this deficiency in forthcoming research.

\subsection{Utility effects}
\label{exp:utility}

\begin{figure}[htb]
    \centering
    \includegraphics[width=\textwidth]{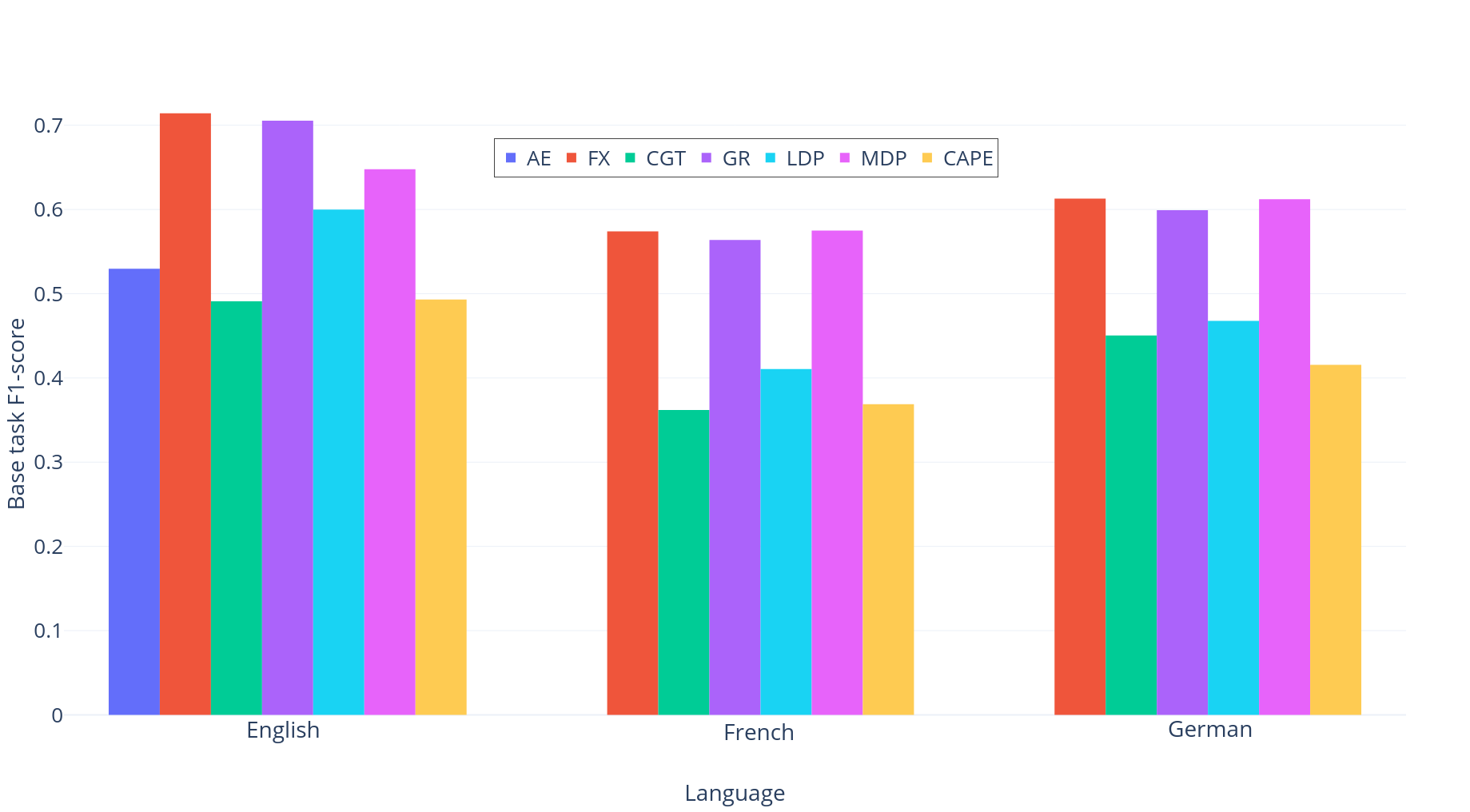}
    \caption{Utility results for English, French, and German models.}
    \label{fig:utility_1}
\end{figure}

\begin{figure}[htb]
    \centering
    \includegraphics[width=\textwidth]{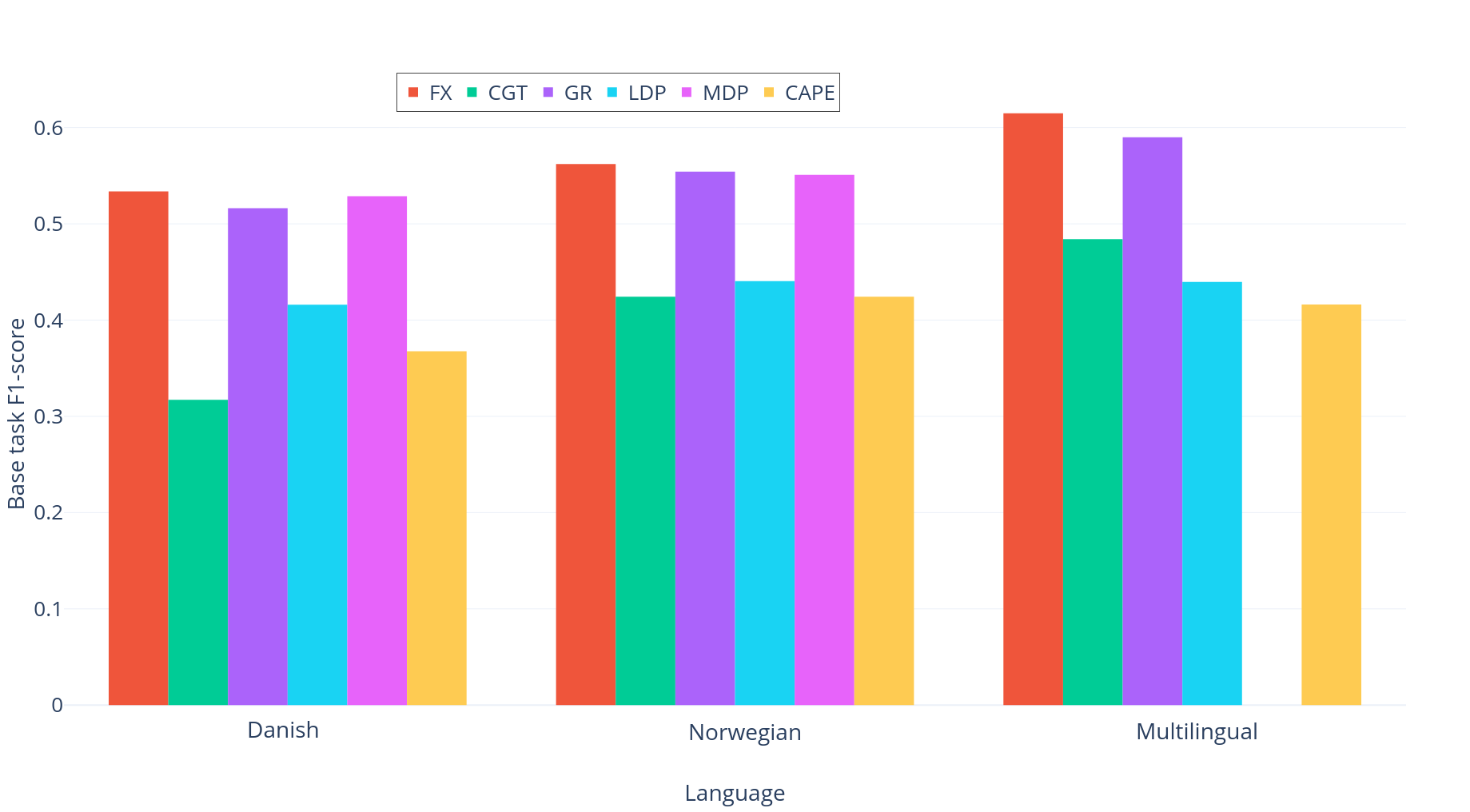}
    \caption{Utility results for Danish, Norwegian, and Multilingual models.}
    \label{fig:utility_2}
\end{figure}

We present here the outcome of our utility testing, where the effect of increasing privacy on the performance of our model is measured as a function of the performance of the base task classifier---as the level of privacy rises, we would expect to see a concomitant drop in utility as a trade off \cite{Li2009obfuscation, Rastogi2007}. We aggregrate the results across each language and privacy method, averaging the F1-score over all runs, including every tested language model/embedding set. Table \ref{tab:utility_results} shows the results, which are also presented in Figures \ref{fig:utility_1} and \ref{fig:utility_2}.

\begin{table}[ht]
    \centering
    \footnotesize
    \begin{tabular}{l|c}
         Model & Base task F1-score \\
         \midrule
         \textbf{English} \\
         AE & 0.530 \\
         FX & 0.714 \\
         CGT & 0.491 \\
         GR & 0.705 \\
         LDP & 0.600 \\
         MDP & 	0.648 \\
         CAPE & 0.493 \\
         \midrule
         \textbf{French} \\
         FX & 0.574 \\
         CGT & 0.362 \\
         GR & 0.564 \\
         LDP & 0.411 \\
         MDP & 	0.575 \\
         CAPE & 0.369 \\
         \midrule
         \textbf{German} \\
         FX & 0.613 \\
         CGT & 	0.450 \\
         GR & 0.599 \\
         LDP & 	0.468 \\
         MDP & 0.612 \\
         CAPE & 0.416 \\
    \end{tabular}
    \quad
    \begin{tabular}{l|c}
         Model & Base task F1-score \\
         \midrule
         \textbf{Danish} \\
         FX & 0.534 \\
         CGT & 0.317 \\
         GR & 0.516 \\
         LDP & 0.416 \\
         MDP & 0.529 \\
         CAPE & 0.368 \\
         \midrule
         \textbf{Norwegian} \\
         FX & 0.562 \\
         CGT & 0.424 \\
         GR & 0.554 \\
         LDP & 0.441 \\
         MDP & 0.551 \\
         CAPE & 0.424 \\
         \midrule
         \textbf{Multilingual} \\
         FX & 0.615 \\
         CGT & 0.484 \\
         GR & 0.590 \\
         LDP & 0.440 \\
         CAPE & 0.416 \\
    \end{tabular}
    \caption{Base task performance F1-score per model per language, averaged across all runs.}
    \label{tab:utility_results}
\end{table}

We can make some initial determinations from an examination of the data:

\begin{itemize}
    \item \textbf{Utility outcomes are better for higher-resourced languages.} Averaged across all models, base task performance for English outpaces all other language groups, exhibiting a maximum $\approx15\%$ absolute lead over the Danish group and minimum $\approx~9\%$ absolute over the multilingual group. Norwegian and Danish results are also lower in absolute terms than French and German, which gives weight to our intuition that this finding is relative to the degree of pre-training carried out and the extent of the training corpus.
    \item \textbf{Gradient reversal and metric differential privacy exhibit relatively fewer negative utility effects.} GR and MDP methods both exhibit only around $\approx 2\%$ drop in base performance in absolute terms, which dependent on the task in question may represent a negligible downstream effect. This finding is interesting given the divergent levels of privacy offered by the respective methods: in English results for instance, GR exhibits a re-identification performance reduction of around $\approx 13\%$ absolute. Given that MDP is a general privatisation measure and does not require the introduction of a known set of demographic annotations, this is perhaps an explainable phenomenon. However, additional testing could reveal the extent of the general privacy guarantee; for more, see Section \ref{sec:further_work}.
    \item \textbf{Local differential privacy introduces a high level of utility perturbance at a fixed privacy budget.} Both systems that rely on additive noise, LDP and CAPE, display a similar level of utility degradation at $\approx 20\%$ absolute penalty, higher than the other systems addressed in our testing. This is to be expected, since we have set a very restrictive privacy budget ($\epsilon = 0.1$) to achieve very high levels of privacy risk reduction in our experimentation. A more permissive set of parameters can help mitigate the utility effect, as discussed in Section \ref{sec:noise}. It must be noted however that the $\epsilon$ variable does not express a general budget factor that is interpretable across all potential tasks; the budget determines the level of additional noise \textit{in proportion} to the sensitivity of queries made during computation, so the correct value must be established for each task individually.
\end{itemize}

\section{Discussion}
\label{sec:discussion}

This section will provide some further context and analysis for our results in the previous experiments, including the impact of some potential confounding factors on the level of privacy provided by our suggested systems. 

\subsection{Assessing embedding deviation}

During our experiments, the question occurred of how to adjudge the potential effects on privacy and utility of preservation techniques without the investment of extensive empirical research. Here, we propose to apply any potential technique which involves altering the content of embeddings to a subset of the dataset, then projecting the vectors into a lower-dimensional space where large-scale effects may be more obvious.

To test this, we gather a slice consisting of the first 10,000 rows in our English language corpus, and extract the pre-trained embeddings from the \textit{bert-base-cased} model. For comparison, we extract the same set of sequences from our metric differential privacy model, and from our local differentially private model. We present the average Euclidean and Cosine distance metrics (calculated for each paired row) in Table \ref{tab:embed_dist}.

\begin{table}[ht]
    \footnotesize
    \centering
    \begin{tabular}{l|cc}
         & Euclidean & Cosine \\
         \midrule
        LDP & 387.987 & 0.987 \\
        MDP & 6.017 & 0.168 \\
    \end{tabular}
    \caption{Average distance metrics for English text sequences paired with equivalents produced by LDP and MDP models.}
    \label{tab:embed_dist}
\end{table}

\begin{figure}[ht]
    \centering
    \includegraphics[width=0.7\textwidth]{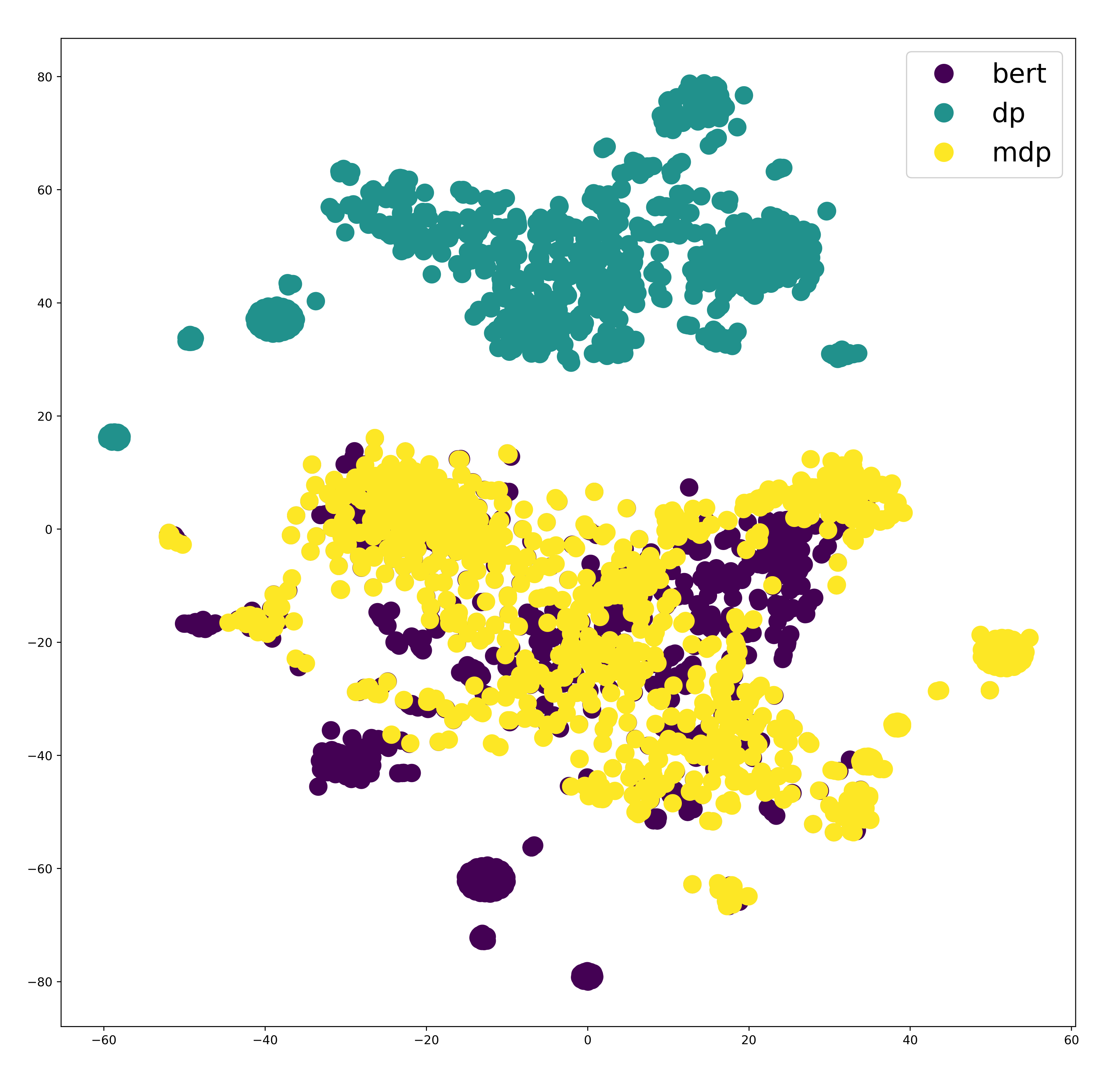}
    \caption{Plot of embedding vectors for English text sequences produced by the 'bert-base-cased' model, and their local DP and metric DP equivalents. Dimensionality reduced using t-SNE.}
    \label{fig:embed_plot}
\end{figure}

Our empirical results demonstrate the distinct effect of the differential privacy process, but the outcome may be easier to distinguish in a visual form. In order to project our dataset slice into a two-dimensional space, we adopt the t-SNE dimensionality reduction process \cite{maaten_visualizing_2008}, which attempts to minimise the KL-divergence between a similarity distribution over high-dimensional objects and a corresponding low-dimensional representation.

The results of the t-SNE dimensionality reduction process can be seen in Figure \ref{fig:embed_plot}, with the original embedding compared to the equivalent sequence with local and metric DP applied. We observe that the outcome of the dimensionality leads to a linearly-separable grouping of local DP sequences, while the original and MDP results are far more contiguous. Intuitively, this supports our earlier results - the LDP-perturbed embeddings are less representative of the original content of the text sequence, and so will exhibit less effective task performance. 

However, the distribution of points does not seem to capture the dynamics evident in our privacy experiments, which indicates the potential for this approach to indicate potential detriments to task performance, in the event of deciding between competing privacy technologies.

\subsection{Analysing the privacy/utility tradeoff}
\label{sec:noise}

In proposing the CAPE model, we proffer the potential for tuning the privacy parameters of the model to account for a range of possible privacy vs. utility scenarios. In order to determine the effect of adjusting these parameters, we carried out a number of experiments over a range of sensible values drawn from the existing literature: in terms of the weight given to the contribution of the adversarial learning system in the model's overall objective function denoted by the variable $\lambda$, we refer to the values established in \citet{ganin_domain-adversarial_2016}, while our range of values for the privacy budget parameter $\epsilon$, which controls the addition of noise under the local differential privacy regime, follows the example of \citet{lyu_differentially_2020}. 

We carry out each run of the CAPE model over the English instances from our dataset, using the \textit{bert-base-cased} embedding set and the \textit{gender} demographic attribute. Results for both the basic review score prediction task and simulated attacker task are presented in Table \ref{tab:noise_tune_cape}.

\begin{table}[ht]
    \footnotesize
    \centering
    \begin{tabular}{ll|cc}
         $\lambda$ & $\epsilon$ & Base task F1 & Attacker F1 \\
         \midrule
        0.1 & 0.01 & 0.449 & 0.375 \\
        0.1 & 0.1 & 0.532 & 0.375 \\
        0.1 & 0.5 & 0.705 & 0.375 \\
        0.1 & 1 & 0.701 & 0.375 \\
        0.1 & 5 & 0.708 & 0.375 \\
        0.1 & 10 & 0.693 & 0.375 \\
        0.1 & 20 & 0.713 & 0.375 \\
        0.1 & 50 & 0.696 & 0.375 \\
        0.1 & 100 & 0.693 & 0.375 \\
        0.5 & 0.01 & 0.449 & 0.375 \\
        0.5 & 0.1 & 0.460 & 0.375 \\
        0.5 & 0.5 & 0.701 & 0.375 \\
        0.5 & 1 & 0.654 & 0.375 \\
        0.5 & 5 & 0.683 & 0.375 \\
        0.5 & 10 & 0.716 & 0.375 \\
        0.5 & 20 & 0.707 & 0.375 \\
        0.5 & 50 & 0.712 & 0.375 \\
        0.5 & 100 & 0.652 & 0.375 \\
        1 & 0.01 & 0.016 & 0.375 \\
        1 & 0.1 & 0.526 & 0.375 \\
        1 & 0.5 & 0.703 & 0.375 \\
        1 & 1 & 0.685 & 0.378 \\
        1 & 5 & 0.711 & 0.375 \\
    \end{tabular}
    \quad
    \begin{tabular}{ll|cc}
        $\lambda$ & $\epsilon$ & Base task F1 & Attacker F1 \\
         \midrule
         1 & 10 & 0.697 & 0.375 \\
        1 & 20 & 0.704 & 0.375 \\
        1 & 50 & 0.680 & 0.375 \\
        1 & 100 & 0.708 & 0.375 \\
        1.5 & 0.01 & 0.449 & 0.294 \\
        1.5 & 0.1 & 0.449 & 0.375 \\
        1.5 & 0.5 & 0.703 & 0.375 \\
        1.5 & 1 & 0.670 & 0.375 \\
        1.5 & 5 & 0.709 & 0.375 \\
        1.5 & 10 & 0.707 & 0.375 \\
        1.5 & 20 & 0.697 & 0.375 \\
        1.5 & 50 & 0.712 & 0.375 \\
        1.5 & 100 & 0.709 & 0.375 \\
        2 & 0.01 & 0.449 & 0.375 \\
        2 & 0.1 & 0.449 & 0.375 \\
        2 & 0.5 & 0.635 & 0.375 \\
        2 & 1 & 0.515 & 0.375 \\
        2 & 5 & 0.567 & 0.375 \\
        2 & 10 & 0.532 & 0.375 \\
        2 & 20 & 0.449 & 0.294 \\
        2 & 50 & 0.685 & 0.375 \\
        2 & 100 & 0.549 & 0.375 \\
    \end{tabular}
    \caption{F1-score for base and simulated attacker tasks for CAPE model over a range of privacy budgets ($\epsilon$) and adversarial weight parameter values ($\lambda$). }
    \label{tab:noise_tune_cape}
\end{table}

\begin{figure}[ht]
    \centering
    \includegraphics[width=0.9\textwidth]{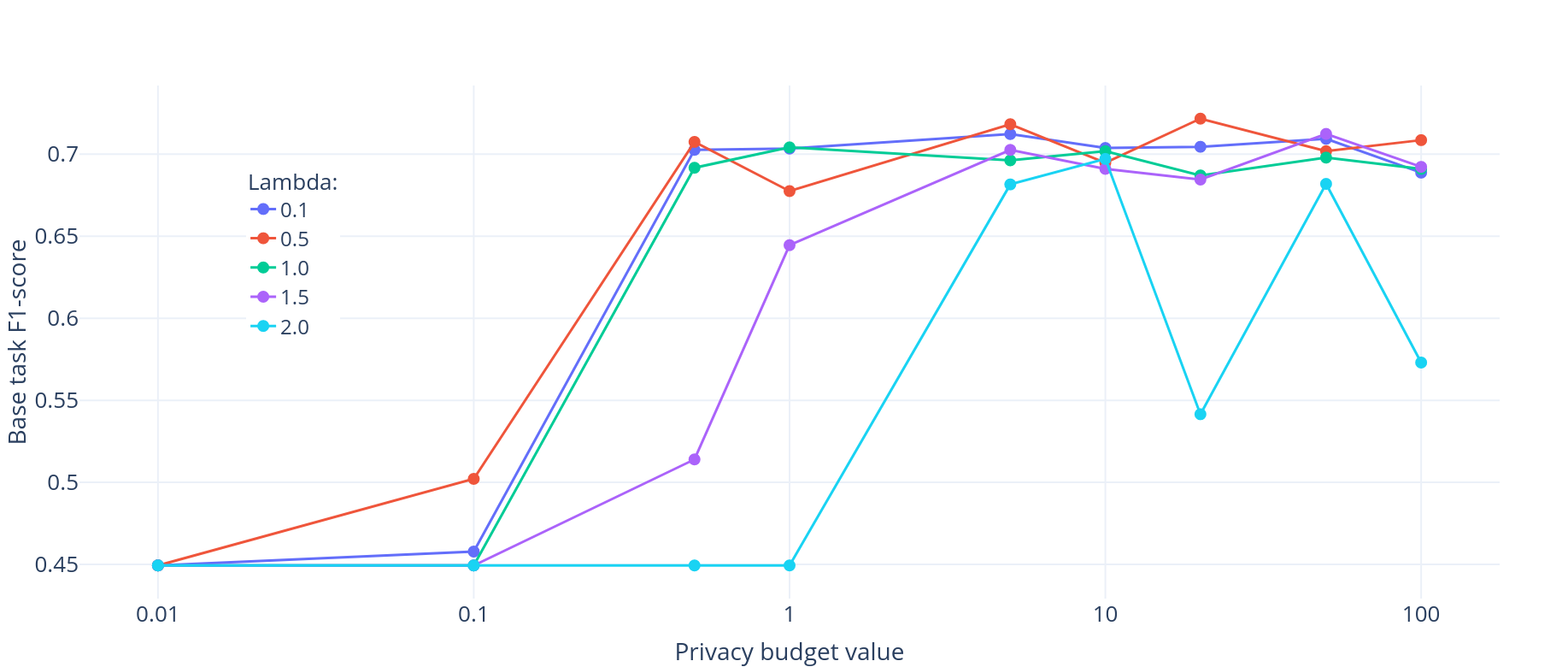}
    \caption{F1-score for basic review score prediction task for the CAPE model over a range of privacy budgets ($\epsilon$), grouped by adversarial weight parameter ($\lambda$). Higher values on the y-axis represent better task performance.}
    \label{fig:noise_tune_base}
\end{figure}

\begin{figure}[ht]
    \centering
    \includegraphics[width=0.9\textwidth]{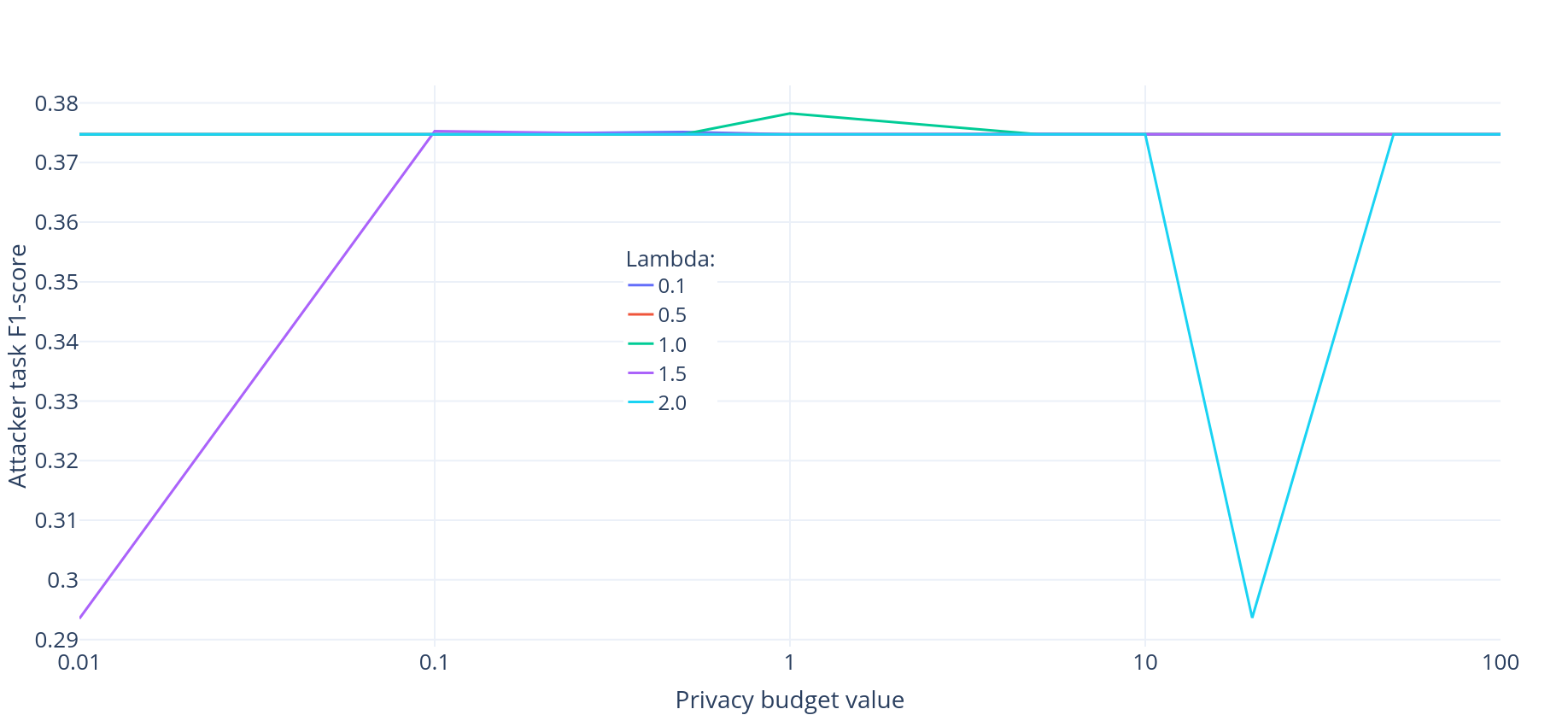}
    \caption{F1-score for simulated attacker task for the CAPE model over a range of privacy budgets ($\epsilon$), grouped by adversarial weight parameter ($\lambda$). Lower values on the y-axis represent better privacy preservation.}
    \label{fig:noise_tune_atk}
\end{figure}

We can immediately determine that base task performance increases broadly in proportion to the size of the privacy parameter $\epsilon$, as was expected, although we can also see that increasing the weighting of the adversarial training objective by raising the value of $\lambda$ above 1.0 tends to negatively affect performance - catastrophically so at the more permissive end of the privacy scale, as can be seen more clearly in Figure \ref{fig:noise_tune_base}. 

However, the more interesting result is displayed in the plot of privacy outcomes for the same set of experiments, shown in Figure \ref{fig:noise_tune_atk}. We can see here that the level of the privacy budget makes a relatively limited impact on the level of information leakage. Note here that we are testing only a single variable, the value of which is used during our adversarial training phase, rather than information that is completely unavailable to the algorithm.

\begin{figure}[ht]
    \centering
    \includegraphics[width=\textwidth]{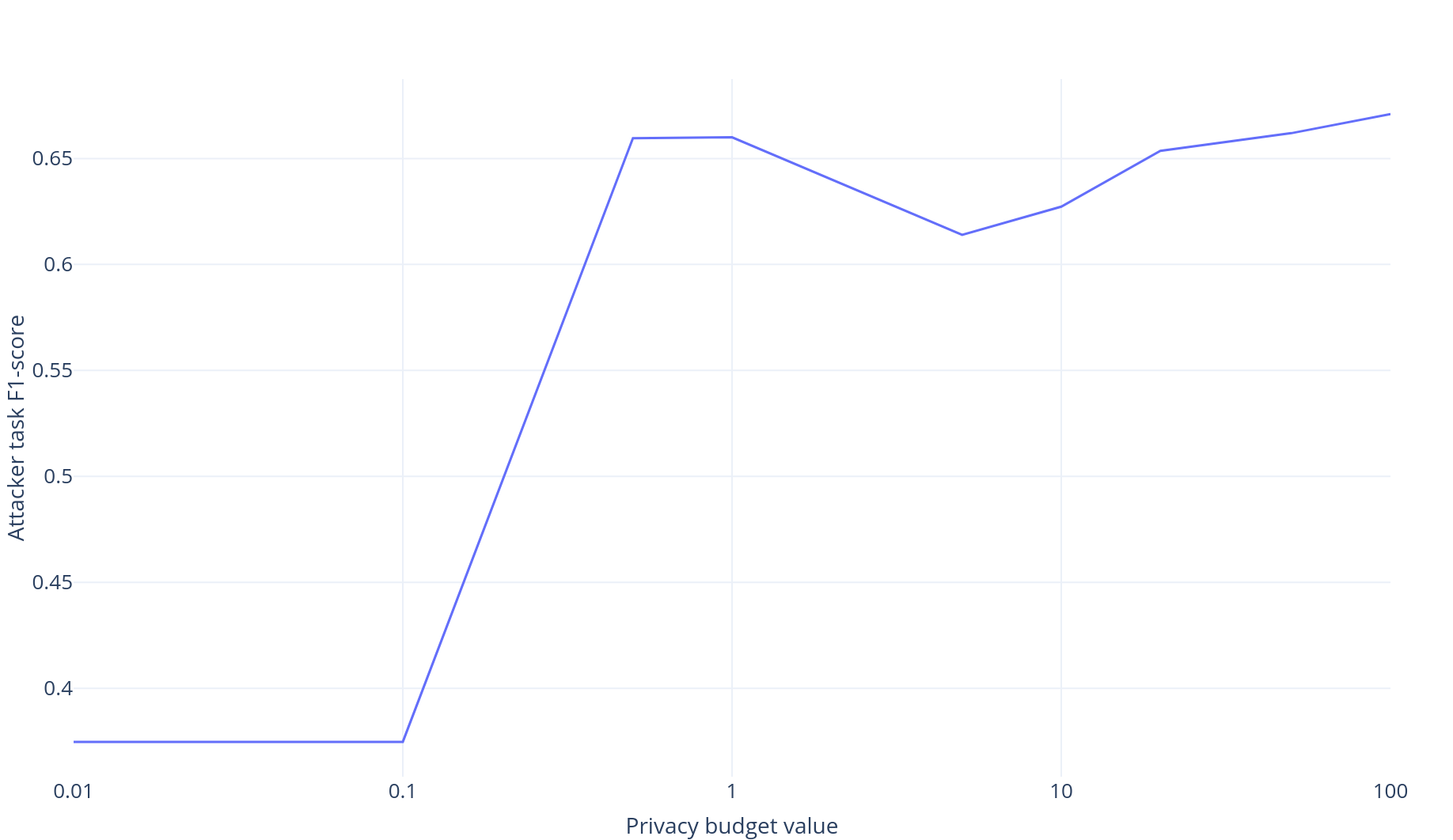}
    \caption{F1-score for simulated attacker task for LDP model over a range of privacy budgets ($\epsilon$). Lower values on the y-axis represent better privacy preservation.}
    \label{fig:noise_tune_ldp}
\end{figure}

This finding is of immediate interest, since if raising the privacy budget allows us to achieve better utility outcomes without sacrificing privacy on the identified demographic variables, then this may go some way towards mitigating the negative impact of differential privacy in NLP task settings. To determine how much of this retained privacy performance is due to the hybrid nature of the model, we repeated the experiments with our local DP-only model using the same set of parameters, achieving the results seen in Table \ref{tab:noise_tune_ldp} and Figure \ref{fig:noise_tune_ldp}.

\begin{table}[ht]
    \small
    \centering
    \begin{tabular}{l|c}
        Privacy budget & Attacker F1 \\
        \midrule
        0.01 & 0.375 \\
        0.1 & 0.375 \\
        0.5 & 0.660 \\
        1 & 0.660 \\
        5 & 0.614 \\
        10 & 0.627 \\
        20 & 0.654 \\
        50 & 0.662 \\
        100 & 0.671 \\
    \end{tabular}
    \caption{F1-score for simulated attacker task for LDP model over a range of privacy budgets ($\epsilon$).}
    \label{tab:noise_tune_ldp}
\end{table}

These results provide an interesting counterpoint to the CAPE model: it appears obvious that at higher---and therefore more permissive---privacy budget values, the protection provided by the adversarial learning objective simply supersedes the LDP privacy mechanism, which would otherwise fail to obscure the information leakage from the sample text. This implies that hybrid mechanisms can be relied on to ensure a minimum level of privacy regardless of privacy budget, as long as the demographic variables are pre-annotated and available at training time.

\subsection{Increasing model complexity}

Finally, we assess here the impact of increasing classifier complexity and depth on privacy outcomes, both those from non-private encodings and from those privatised using our CAPE model. Since we would expect a deeper classifier network to be prima facie better at distinguishing demographic signal from the noise of everything else encoded in a pre-trained sequence embedding, we would expect a more complex simulated attacker network to therefore exhibit higher information leakage in a non-private setting, and potentially a higher relative drop in attacker performance when privacy preservation is applied. 

We carry out a set of experiments across a range of classifier setups for both private and non-private scenarios, using the English language splits from our dataset and obtaining the embeddings from the 'bert-base-cased' language models. We adapt our attacker network by altering the 'depth'---that being the number of layers in the classifier---and also the 'width'---that being the number of units in each layer.

\begin{table}[ht]
    \centering
    \footnotesize
    \begin{tabular}{ll|ccc}
         Depth & Width & Non-private & CAPE & Change \\
         \midrule
         1 & 50 & 0.681 & 0.394 & -42\% \\
        1 & 100 & 0.666 & 0.375 & -44\% \\
        1 & 200 & 0.663 & 0.375 & -43\% \\
        1 & 500 & 0.680 & 0.375 & -45\% \\
        2 & 50 & 0.669 & 0.294 & -56\% \\
        2 & 100 & 0.670 & 0.293 & -56\% \\
        2 & 200 & 0.671 & 0.375 & -44\% \\
        2 & 500 & 0.652 & 0.375 & -43\% \\
        3 & 50 & 0.678 & 0.375 & -45\% \\
        3 & 100 & 0.684 & 0.375 & -45\% \\
        3 & 200 & 0.678 & 0.375 & -45\% \\
        3 & 500 & 0.647 & 0.375 & -42\% \\
        4 & 50 & 0.685 & 0.375 & -45\% \\
        4 & 100 & 0.675 & 0.375 & -44\% \\
        4 & 200 & 0.685 & 0.375 & -45\% \\
        4 & 500 & 0.684 & 0.375 & -45\% \\
        5 & 50 & 0.674 & 0.375 & -44\% \\
        5 & 100 & 0.646 & 0.375 & -42\% \\
        5 & 200 & 0.656 & 0.375 & -43\% \\
        5 & 500 & 0.673 & 0.375 & -44\% \\
    \end{tabular}
    \caption{Attacker performance as F1-score for non-private and CAPE models. Depth refers to the number of layers in the classifier network, width to the number of units in each layer.}
    \label{tab:model_depth}
\end{table}

\begin{figure}[htb]
    \centering
    \includegraphics[width=\textwidth]{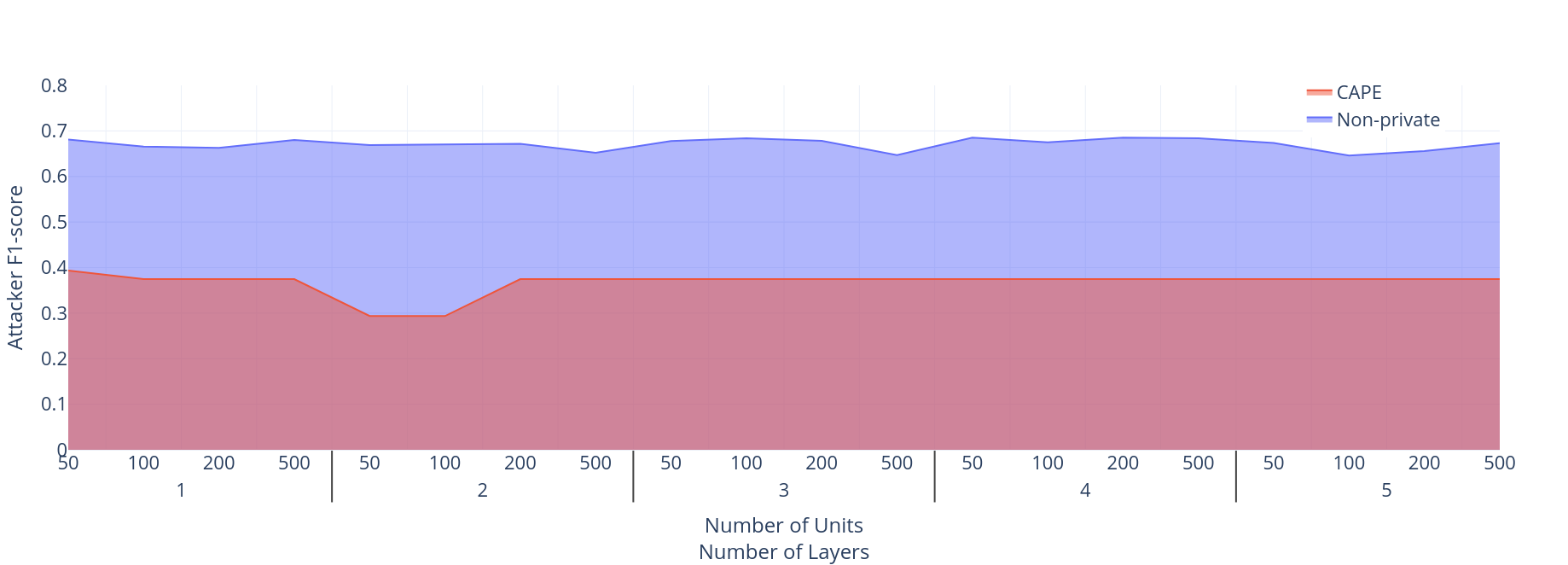}
    \caption{Attacker performance over range of depth/width values.}
    \label{fig:model_depth}
\end{figure}

Results are presented in Table \ref{tab:model_depth}, as the F1-score representing the attacker performance for both the non-private and private networks, with the relative change between the settings also reported. Interestingly, the results do not bear out our assumptions; looking at Figure \ref{fig:model_depth}, it is clear that the relative performance of our adversary network does not vary significantly with the complexity of the classifier layers. This may indicate that the attacker network topology is not germane to the detection of the relatively robust demographic signal from the pre-trained embedding, although this would require additional experimentation with alternate network components to judiciously assess.

\section{Conclusion}
\label{sec:conclusion}

In this work, we have presented an initial set of results that we believe show an indication the extent of the information leakage issue for language models; beyond that, we believe we have presented some indicative results for practical solutions that may help ameliorate the issue---a concrete necessity given that many governments are evincing interest in regulating the practice \cite{aho_beyond_2020, minssen_regulatory_2020}. We have not only proposed a simple framework for assessing the practical consequences of implementing various privacy preserving mechanisms, but we have also extended that toolset to cover multiple potential areas of interest: alternate languages, language models, and types of private information. We have covered in our language model selection a broad section of popular models which we consider representative \cite{qiu_pre-trained_2020}, however given the speed at which new models are being derived, we would expect this list of change radically in a relatively short space of time.

In performing our evaluation, we have been able to rely to a large extent on the valuable contribution of the Trustpilot dataset, which as a large multilingual resource with reliable annotation in multiple demographic categories provides an ideal test case for our test protocol. However, the availability of such datasets is severely limited, particularly when it comes to the inclusion of language sets under-represented in the field at large \cite{benjamin_hard_2018, mazarura_semantic_2019}. Without high-quality resources with useful indicative variable annotations in the language in question, large scale cross-comparison of privacy mechanisms could prove difficult in the extreme. 

Based on our experimental evidence, we established the following set of axioms that may provide a useful resource for further research in the field:
 
 \begin{itemize}
     \item \textbf{All pre-trained language models leak demographic data.} We were able to extract useful leaked information from all of the models during our testing, and convert that into inferences about user demographics that they could reasonably expect to remain private.
     \item \textbf{The complexity of the language model, and how much data was used to train it, matters.} More complex language models, and those trained with a more extensive corpus of data, were more prone to leak information - and vice versa.
     \item \textbf{Classifier complexity is not highly linked to attacker performance.} Our tests found no significant correlation between attacker performance and the depth of the classifier network.
     \item \textbf{Local differential privacy is highly deleterious to model utility.} Applying noise across embeddings at the level of privacy budget tested necessitated a significant cost to base task performance in our testing, which we can partially control for in hybrid models with laxer budgets.
     \item \textbf{Metric differential privacy shows much more moderate impact on performance.} The simple Euclidean metric differential privacy implemented here exhibited far lower utility costs, while providing somewhat less impressive privacy results.
 \end{itemize}

As LLMs continue to become available at greater scale and complexity---during the writing of this work, the 175 billion parameter GPT-3 model \cite{brown_language_2020} was made public---we believe this research, and the objective of achieving measurable and empirically verifiable privacy preservation more broadly, only becomes more necessary. Without a commonly agreed framework for making decisions on the acceptable level of information leakage, we run the risk of undermining user trust.

\subsection{Further Work}
\label{sec:further_work}

Finally, we would point to the following fruitful lines of inquiry we noted during the preparation of this work:

\begin{itemize}
    \item In our experiments, we assessed the privacy performance of differentially private and hybrid models using the same set of private variables as we included in the set seen by all the models at training time. In order to test the general level of privacy provided by DP on unseen variables over adversarial learning, it would be desirable to engineer a scenario where we introduce new variables during evaluation that are not included in training.
    \item One potential avenue of research we did not pursue relates to the impact of representation dimensionality on information leakage; if it is the case that higher-dimensional encodings capture a more nuanced set of interactions between individual tokens within a sequence, then producing lower-dimensional representations with the same model may provide a shortcut to leaking less additional information. This would require retraining a chosen language model several times.
    \item The question of privacy attacks other than re-identification of personal attributes remains unaddressed by our work. One immediate extension we envisage is to also produce results for membership inference attacks: a form of attack where an adversary uses model outputs to infer whether an individual's data was used in the training set for a model \cite{shokri_membership_2017, leino_stolen_2020}.
    \item In this work, we deal with perhaps the most simplistic implementation of metric differential privacy, projecting each embedding in Euclidean space. However, multiple alternate formulations that use different metrics have been proposed \cite{feyisetan_leveraging_2019, xu_differentially_2020, carvalho_tem_2021}. Evaluating empirically the distinct outcomes of each distance metric would be a worthy endeavour.
\end{itemize}

Beyond these immediate concerns, it is incumbent on us to point out that while language model training and privacy preservation remain separate concerns, including a high level of user protection will represent an additional cost in terms of implementation and research, which may not be bearable by lower-resourced users of such models. We propose that the wholesale importation of privacy preservation techniques into language model production---for instance, by the addition of additional objectives which penalise attention to potentially revealing tokens---could help address the information leakage issue at source.

\clearpage

\appendix
\appendixsection{Dataset Description}
\label{app:dataset}

Since a long-form description of the \citet{Hovy} dataset as proscribed by \citet{bender_data_2018} is not available, we have prepared a description of the characteristics of the dataset to aid in recognising the generalisation potential of our research.

\textbf{Curation Rationale}
\label{sec:curation}
In order to provide a large scale textual resource with high coverage of demographic annotations for study \cite{Hovy} crawled the Trustpilot website, extracting publicly available review content as JSON-formatted objects. Data in the released set is restricted to those countries with more than 250,000 users at the time of collection in 2015: Denmark, France, Germany, the United Kingdom, and the United States. Collection was restricted to reviews posted at most seven years prior to the date of collection.

The collected set is augmented in two ways. Firstly, when gender information is not supplied by the reviewer, the researchers attempt to add gender by reference to the existing distribution of first names and genders. If a particular first name appears more than 3 times in the un-augmented set and is correlated with one gender at least 95\% of the time, that gender is propagated to all other occurrences of that first name without an attached gender. Secondly, the researchers use the Geonames database to add latitude and longitude information to the free-text location field as reported by the reviewer. In occasions where the canonical location is indeterminate, such as when two towns share the same name, the researchers select the largest town of that name in the selected country. 

\textbf{Language Variety}
Languages represented within the dataset with more than 100 instances are listed in Table \ref{tab:langs} as represented by ISO639-1 codes. Regional variations are not listed, but for English are likely to include both British English (en-UK) and US English (en-US).

\begin{table}[h]
    \centering
    \small
    \begin{tabular}{c|c}
    \toprule
        Language Code & Instances  \\
        \midrule
        en & 1567424\\
        da & 935629\\
        fr & 430103\\
        no & 184126\\
        de & 176115\\
        nb & 24767\\
        nl & 8102\\
        sv & 7823\\
        es & 7691\\
        it & 5891\\
        ca & 5449\\
        pl & 2469\\
        wa & 2191\\
        pt & 1948\\
        nn & 1946\\
        eo & 1275\\
        mt & 1077\\
        fi & 1051\\
        et & 964\\
    \end{tabular}
    \quad
    \begin{tabular}{c|c}
    \toprule
        Language Code & Instances \\
        \midrule
        oc & 907\\
        ro & 582\\
        sl & 545\\
        ms & 527\\
        id & 415\\
        ht & 371\\
        hu & 344\\
        tr & 330\\
        lt & 306\\
        cy & 254\\
        af & 244\\
        eu & 236\\
        tl & 222\\
        zh & 192\\
        am & 180\\
        br & 162\\
        cs & 152\\
        ug & 121\\
        sk & 110\\
    \end{tabular}
    \caption{Language tag and no. of examples in dataset}
    \label{tab:langs}
\end{table}

\textbf{Speaker Demographic}
Speakers were not directly invited to participate in this set and a full demographic overview is not available. No specific information about the users' income, social or economic status is available explicitly within the dataset.

However, the researchers do have access to the self-reported demographic attributes of age, gender and location. Table \ref{tab:dems} shows the proportion of users in each country for which those attributes are available.

\begin{table}[h]
    \centering
    \small
    \begin{tabular}{l|ccccc}
    \toprule
         & Users & Age & Gender & Location & All  \\
         \midrule
        United Kingdom & 1,424k & 7\% & 62\% & 5\% & 4\%\\
        France & 741k & 3\% & 53\% & 2\% & 1\%\\
        Denmark & 671k & 23\% & 87\% & 17\% & 16\%\\
        United States & 648k & 8\% & 59\% & 7\% & 4\%\\
        Netherlands & 592k & 9\% & 39\% & 7\% & 5\%\\
        Germany & 329k & 8\% & 47\% & 6\% & 4\%\\
        Sweden & 170k & 5\% & 64\% & 4\% & 3\%\\
        Italy & 132k & 10\% & 61\% & 8\% & 6\%\\
        Spain & 56k & 6\% & 37\% & 5\% & 3\%\\
        Norway & 51k & 5\% & 50\% & 4\% & 3\%\\
        Belgium & 36k & 13\% & 42\% & 11\% & 8\%\\
        Australia & 31k & 8\% & 36\% & 7\% & 5\%\\
        Finland & 16k & 6\% & 36\% & 5\% & 3\%\\
        Austria & 15k & 10\% & 43\% & 7\% & 5\%\\
        Switzerland & 14k & 8\% & 41\% & 7\% & 4\%\\
        Canada & 12k & 10\% & 19\% & 9\% & 4\%\\
        Ireland & 12k & 8\% & 30\% & 7\% & 4\%
    \end{tabular}
    \caption{Proportion of users with available demographic attributes \cite{Hovy}}
    \label{tab:dems}
\end{table}

The mean and median ages for users in the sets for each country are given in Table \ref{tab:ages}, in comparison to the official figures for each country as reported in the CIA World Factbook \cite{central_intelligence_agency_median_2021}.

\begin{table}[h]
    \centering
    \small
    \begin{tabular}{l|cc|c||cc|c}
    \toprule
        & \multicolumn{3}{c}{Women} & \multicolumn{3}{c}{Men}\\
        & mean & median & off. median & mean & median & off. median\\
        \midrule
        Denmark & 38.80 & 38 & 41.6 & 39.07 & 38 & 39.8\\
        France & 42.03 & 41 & 41.2 & 41.92 & 41 & 38.2\\
        Germany & 40.64 & 40 & 45.3 & 38.97 & 38 & 42.3\\
        United Kingdom & 44.51 & 45 & 41.5 & 43.87 & 43 & 39.4\\
        United States & 40.79 & 40 & 38.1 & 36.70 & 33 & 35.5\\
    \end{tabular}
    \caption{Average age per country \cite{Hovy}}
    \label{tab:ages}
\end{table}

\textbf{Annotator Demographic}
The dataset is not substantially annotated except where demographic attributes have been inferred using the augmentation techniques described in Section \ref{sec:curation}.

\textbf{Speech Situation}
All reviews were posted to the Trustpilot website between 2008 and 2015. Messages represent extemporaneous literature describing the experience of users with various businesses, services, products, and other agencies. The public nature these communications, along with the associated star ratings, indicates that these were regarded by the user as intended for public consumption, with the intended audience presumably prospective users of the target business or service. A secondary target of the speech may also have been the operators of such businesses. 

\begin{figure}[ht]
    \centering
    \includegraphics[width=\textwidth]{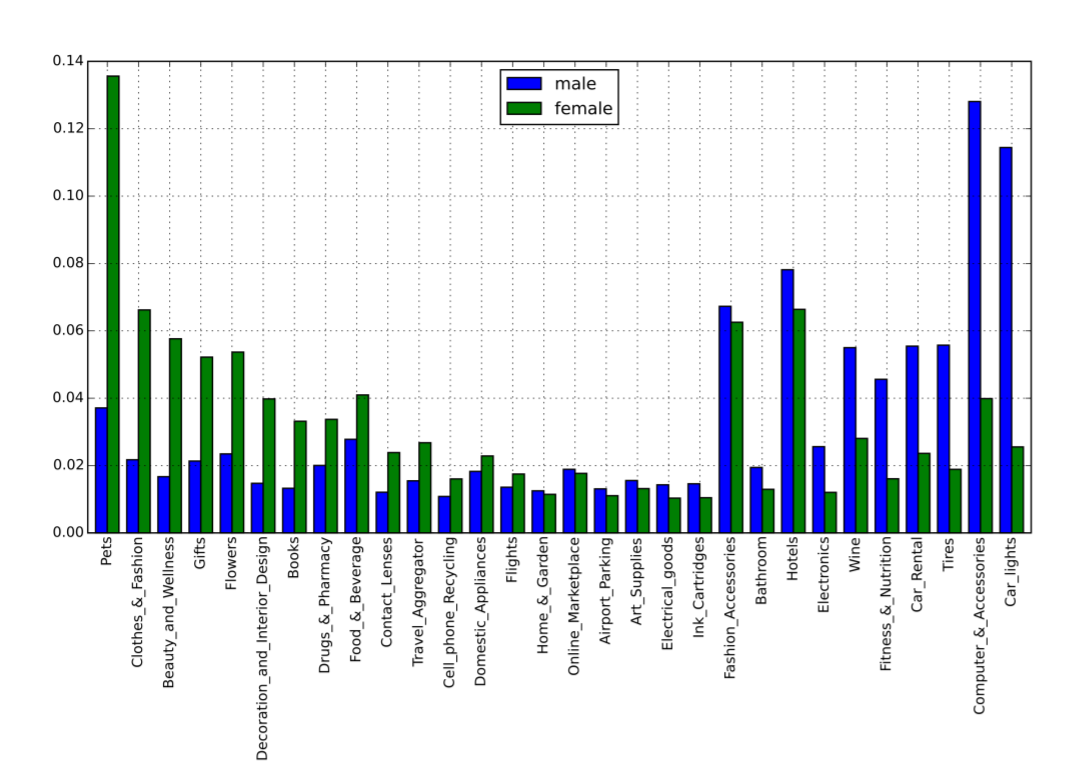}
    \caption{Most common review categories by gender \cite{Hovy}}
    \label{fig:topic}
\end{figure}

\textbf{Text Characteristics}
The majority of texts in this dataset concern reviews of products and retail services such as parking charges or home decoration, rather than other forms of business transaction such as legal or business consultancy. A distribution across the most common categories addressed by the review broken down by gender \citep{Hovy} is illustrated in Figure \ref{fig:topic}.

\clearpage

\appendixsection{Language Models}
\label{app:lang_models}

\begin{table}[ht]
    \centering
    \begin{tabular}{c|c}
         Language & Model \\
         \midrule
         English & bert-base-cased \\
         & bert-base-multilingual-cased \\
         & roberta-base \\
         & albert-base \\
         & gpt2 \\
         \midrule
         French & flaubert/flaubert\_base\_cased \\
         & bert-base-multilingual-cased \\
         & camembert-base \\
         & antoiloui/belgpt2 \\
         \midrule
         German & bert-base-german-cased \\
         & bert-base-multilingual-cased \\
         & T-Systems-onsite/cross-en-de-roberta-sentence-transformer \\
         & dbmdz/german-gpt2 \\
         \midrule
         Danish & Maltehb/danish-bert-botxo \\
         & bert-base-multilingual-cased \\
         & flax-community/roberta-base-danish \\
         \midrule
         Norwegian & ltgoslo/norbert \\
         & bert-base-multilingual-cased \\
         & flax-community/nordic-roberta-wiki \\
         \midrule
         Multilingual & bert-base-multilingual-cased \\
         & paraphrase-xlm-r-multilingual-v1 \\
         & distiluse-base-multilingual-cased-v2 \\
    \end{tabular}
    \caption{Language models used}
    \label{tab:app_lang_models}
\end{table}

\newpage

\starttwocolumn
\bibliography{references}

\end{document}